\documentclass[letterpaper]{article} 
\usepackage{aaai2026}
\usepackage{times}  
\usepackage{helvet}  
\usepackage{courier}  
\usepackage[hyphens]{url}  
\usepackage{graphicx} 
\usepackage{natbib}  
\usepackage{caption} 
\usepackage{algorithm}
\usepackage{algorithmic}

\usepackage{amsmath,amssymb,amsfonts}
\usepackage{booktabs}
\usepackage{multirow}
\usepackage[table]{xcolor}
\usepackage{arydshln}
\usepackage{subcaption}
\usepackage{siunitx}

\usepackage[hyphens]{url}

\usepackage{newfloat}
\usepackage{listings}
\DeclareCaptionStyle{ruled}{labelfont=normalfont,labelsep=colon,strut=off} 
\floatstyle{ruled}
\newfloat{listing}{tb}{lst}{}
\floatname{listing}{Listing}
 \nocopyright 

\usepackage{xspace}
\newcommand{\ourname}{\texttt{CoPLOT}\xspace}


\newif\ifshowappendix
\showappendixtrue

\newif\ifshowchecklist
\showchecklisttrue

\newcommand{\myeq}[1]{Eq.~(\ref{#1})}
\newcommand{\myfig}[1]{Fig.~\ref{#1}}
\newcommand{\mytab}[1]{Tab.~\ref{#1}}

\usepackage{xcolor}

\title{
Beyond BEV: Optimizing Point-Level Tokens for Collaborative Perception
}

\author {
    Yang Li, Quan Yuan, Guiyang Luo, Xiaoyuan Fu,  Rui Pan,  \\
    Yujia Yang,  Congzhang Shao, Yuewen Liu, Jinglin Li
}
\affiliations {
    State Key Laboratory of Networking and Switching Technology, \\
    Beijing University of Posts and Telecommunications, Beijing, China \\
    \{leeyang866, yuanquan, luoguiyang, fuxiaoyuan, panrui805, yangyj2022, shaocongzhang, liuyw, jlli\}@bupt.edu.cn
}

\usepackage{bibentry}

\begin{document}

\maketitle

\begin{abstract}

Collaborative perception allows agents to enhance their perceptual capabilities by exchanging intermediate features. 
Existing methods typically organize these intermediate features as 2D bird’s-eye-view (BEV) representations, which discard critical fine-grained 3D structural cues essential for accurate object recognition and localization.
To this end, we first introduce point-level tokens as intermediate representations for collaborative perception.
However, point-cloud data are inherently unordered, massive, and position-sensitive, making it challenging to produce compact and aligned point-level token sequences that preserve detailed structural information.
Therefore, we present \ourname, a novel \textbf{Co}llaborative perception framework that utilizes \textbf{P}oint-\textbf{L}evel \textbf{O}ptimized \textbf{T}okens. 
It incorporates a point-native processing pipeline, including token reordering, sequence modeling, and multi-agent spatial alignment.
A semantic-aware token reordering module generates adaptive 1D reorderings by leveraging scene-level and token-level semantic information. 
A frequency-enhanced state space model captures long-range sequence dependencies across both spatial and spectral domains, improving the differentiation between foreground tokens and background clutter.
Lastly, a neighbor-to-ego alignment module applies a closed-loop process, combining global agent-level correction with local token-level refinement to mitigate localization noise.
Extensive experiments on both simulated and real-world datasets show that \ourname outperforms state-of-the-art models, with even lower communication and computation overhead.
Code will be available at \url{https://github.com/CheeryLeeyy/CoPLOT}.

\end{abstract}


\section{Introduction}


Collaborative perception enables multiple agents to share complementary sensory information, mitigating the inherent limitations of individual platforms, such as occlusions and constrained sensing ranges, thereby significantly enhancing overall perception capabilities \cite{gaoSurveyCollaborativePerception2024,chenEndtoendAutonomousDriving2024}. Leveraging advancements in multi-agent communication technologies, such as vehicle-to-everything (V2X), collaborative perception has been widely adopted for tasks including 3D object detection, semantic segmentation, 3D occupancy prediction, and trajectory forecasting, achieving notable performance improvements \cite{chenEndtoendAutonomousDriving2024}.

In recent years, substantial research efforts have been dedicated to developing efficient and reliable intermediate fusion methods to facilitate real-world deployment \cite{liuVehicletoeverythingAutonomousDriving2023}. Existing approaches typically adopt a pipeline that generates, transmits, and fuses 2D bird’s-eye-view (BEV) representations \cite{baiSurveyFrameworkCooperative2024}. This strategy strikes an effective balance between perception performance and communication overhead, garnering widespread attention in the field \cite{yazganSurveyIntermediateFusion2024}.

Nevertheless, these approaches typically reduce 3D scene data into 2D feature maps, discarding critical fine-grained structural cues, particularly vertical information, which often correlates with categories, terrain variations, and spatial delineation in the scene \cite{liOccMambaSemanticOccupancy2025}.
Exchanging 3D representations that preserve finer structural details than 2D BEV maps can markedly enhance performance on more complex perception tasks.
Moreover, intermediate fusion pipelines typically maintain dense 2D feature maps throughout the process, even though large portions of these maps correspond to empty or background noise irrelevant to object perception. This results in unnecessary computational and communication costs \cite{xuV2XViTVehicletoeverythingCooperative2022}. Fusion of multi-agent views expands the resulting feature maps, also introducing additional computational overhead. Focusing exclusively on the feature representations of potential objects, independent of the scene’s overall scope, offers a more efficient alternative that substantially reduces this overhead.
These limitations raise a critical research question: \textit{How can we design a collaborative framework that moves beyond BEV representations to support efficient computation and exchange of detailed 3D structural and semantic information?}


Building on these insights, we propose \ourname, a novel \textbf{Co}llaborative framework that leverages \textbf{P}oint-\textbf{L}evel \textbf{O}ptimized \textbf{T}okens as intermediate representations. 
These point-level tokens encapsulate comprehensive 3D structural information and object-related semantic cues, enabling the pipeline to focus computation and communication resources on tokens associated with potential objects.
Our \ourname takes advantage of the strength of state space models (SSMs)  in modeling long-range dependencies with linear computational complexity. However, challenges still remain when directly applying SSMs to large-scale 3D point-cloud scenes. 
First, state space modeling requires reordering 3D point-level tokens into a 1D causal sequence, which introduces strong inductive bias and can impair model performance. Existing reordering schemes, such as raster scans or space-filling curves \cite{shiVmambaIRVisualState2025, liOccMambaSemanticOccupancy2025}, rely on fixed patterns and often require multi-directional passes to capture sufficient object-related semantic information, which increases computational overhead and still fails to preserve spatial proximity effectively in dynamic scenes.
Moreover, the abundance of background tokens in 3D point-cloud scenes, intermingled with sparse foreground objects sharing similar spatial distributions, renders the accurate detection and localization of these objects particularly challenging. 
Furthermore, when fusing tokens from multiple agents, the issue of token misalignment due to location noise is not explicitly addressed.

\begin{figure*}[!htb]
    \centering
    \includegraphics[width=1.005\textwidth]{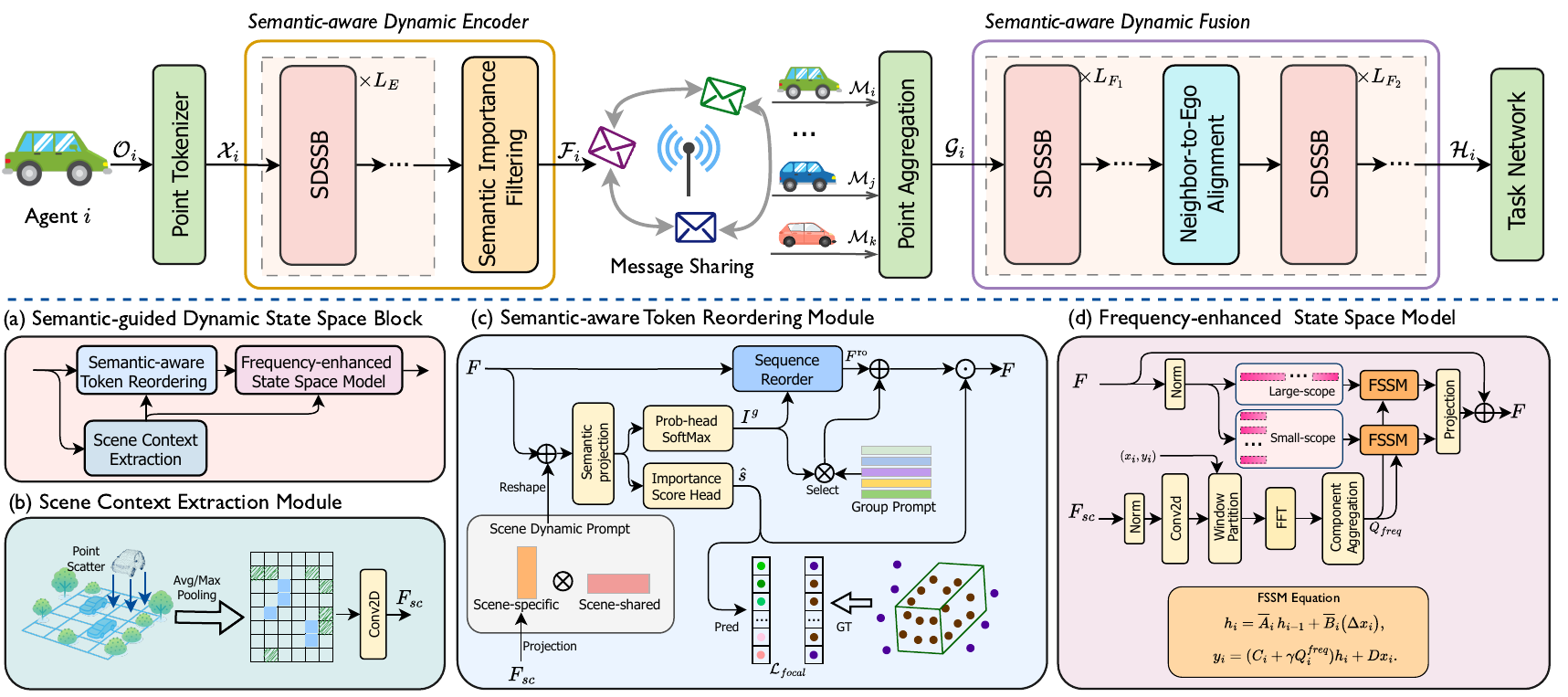}
    \caption{
    The overall architecture of our proposed \ourname, including the submodules: (a) Semantic-guided Dynamic State Space Block (SDSSB),  (b) Scene Context Extraction Module (SCE), (c) Semantic-aware Token Reordering Module (STR), and (d) Frequency-enhanced State Space Model (FSSM).
}
    \label{fig:structure}
\end{figure*}

To this end, we design a tailored point-native processing pipeline for \ourname, as illustrated in \myfig{fig:structure}. The proposed \ourname includes three key components:  
1) A semantic-aware token reordering module that improves the sequence ordering from 3D to 1D. By integrating semantic information from both the global scene level and individual token level, the module precisely analyzes each token’s characteristics and arranges semantically related tokens in close proximity, thereby laying a solid foundation for effective sequence modeling.
2) A frequency-enhanced state space model that further distinguishes the semantic characteristics of point-level tokens and facilitate the selection of object-related key tokens. Drawing inspiration from biological vision systems that employ frequency-selective filters to separate targets from complex backgrounds \cite{acklehFrequencydependentEvolutionPredator2021,zhongDetectingCamouflagedObject2022}, our approach focuses on extracting frequency-domain representations as auxiliary information. These token-specific frequency features are dynamically obtained and integrated into the long-range dependency modeling process, enhancing the model’s capacity to isolate object-relevant cues from background clutter. This mechanism enables the generation of compact point-level token sequences for collaboration, thereby improving both computational and communication efficiency.
3) A neighbor-to-ego alignment module that corrects spatial misalignment between ego and neighbor tokens.
It performs a learnable closed-loop alignment, combining global agent-level correction with local token-level refinement. Tokens adjust their spatial positions based on object-related structural cues while maintaining overall consistency. This explicit realignment enhances the reliability of feature fusion in noisy real-world conditions.

In short, our main contributions are as follows:
\begin{itemize}
\item We introduce \ourname, a novel collaborative framework that operates beyond BEV representations by utilizing point-level optimized tokens, effectively enhancing perception performance while reducing both computational and communication overhead.
\item We develop a tailored point-native processing pipeline for 3D point-level token sequences, with core components comprising a semantic-aware token reordering module, a frequency-enhanced state space model, and a neighbor-to-ego alignment module.
\item We conduct extensive experiments on both simulated and real-world datasets, including OPV2V, V2V4Real, and DAIR-V2X. Results show that our method improves perception performance by up to 10\%, while reducing computational and communication overhead by approximately 80\% and 90\%, respectively.
\end{itemize}

\section{Related Work}

\textbf{Collaborative Perception} enhances accuracy, robustness, and coverage by sharing sensory data across multiple agents \cite{chenEndtoendAutonomousDriving2024,mao3DObjectDetection2023}. Existing methods are commonly categorized into early, late, and intermediate fusion based on the type of shared information. Intermediate fusion, which transmits 2D BEV representations, is widely adopted for its balance between performance and communication efficiency \cite{yazganSurveyIntermediateFusion2024}. Early approaches, such as V2VNet \cite{wangV2VNetVehicletoVehicleCommunication2020}, aimed to improve coordination for joint perception and prediction, while DiscoNet \cite{liLearningDistilledCollaboration2021a} employed knowledge distillation to boost collaboration efficiency. Later methods, including V2X-ViT \cite{xuV2XViTVehicletoeverythingCooperative2022} and CoBEVT \cite{xuCoBEVTCooperativeBirds2022} for cooperative detection and BEV semantic segmentation, capitalized on the strong modeling capacity of Transformers but incurred significant computational overhead. To reduce communication overhead, AttFuse \cite{xuOPV2VOpenBenchmark2022a} introduced channel-wise compression via autoencoders, albeit at the cost of accuracy under high compression. When2com \cite{liuWhen2comMultiAgentPerception2020} and Who2com \cite{liuWho2comCollaborativePerception2020} further enhanced efficiency through scheduling and selective feature sharing. Where2comm \cite{huWhere2commCommunicationEfficientCollaborative2022b} used confidence maps to identify informative spatial regions. How2comm \cite{yangHow2commCommunicationEfficientCollaborationPragmatic2023} and Select2Col \cite{liuSelect2ColLeveragingSpatialtemporal2024} extended this by considering spatio-temporal importance to select optimal collaborators, and CoSDH \cite{xuCoSDHCommunicationefficientCollaborative2025} employed supply-demand aware selection on multi-scale feature maps, preserving accuracy via adaptive filtering. However, these methods rely on 2D feature maps that discard vertical information during height compression. Any further filtering on such representations erodes fine-grained 3D structural detail and undermines the quality of shared features. 
In addition, recent approaches such as CollaMamba \cite{liCollaMambaEfficientCollaborative2024} and CoMamba \cite{liCoMambaRealtimeCooperative2024} adopt efficient emerging state space models to process 2D intermediate features, but they still fail to retain critical 3D structural information.

\textbf{State Space Models} (SSMs) have been widely adopted as competitive backbones in computer vision owing to their high computational efficiency \cite{yaoSelectiveVisualPrompting2025, wangStateSpaceModel2024}, positioning them as strong alternatives to Transformers. However, because SSMs are inherently designed for 1D language sequences, they encounter difficulties when applied to 2D and 3D vision tasks \cite{xuVisualMambaSurvey2024}. To address long-range dependency modeling, Vision Mamba \cite{zhuVisionMambaEfficient2024} and VMamba \cite{liuVMambaVisualState2024a} introduced multi-directional grid scanning. VideoMamba \cite{parkVideoMambaSpatiotemporalSelective2024, liVideoMambaStateSpace2025} extended this approach with spatiotemporal forward and backward scans to model long video sequences. Frequency-based variants such as FreqMamba \cite{zouFreqMambaViewingMamba2024} and DiMSUM \cite{phungDiMSUMDiffusionMamba2024} employ grouped frequency-domain scanning, achieving superior performance in image editing and generation tasks. In the context of 3D point clouds, Mamba3D \cite{hanMamba3DEnhancingLocal2024} uses grouped scans to extract both global and local features, while Voxel Mamba \cite{zhangVoxelMambaGroupfree2024}, OccMamba \cite{liOccMambaSemanticOccupancy2025}, and UniMamba \cite{jinUniMambaUnifiedSpatialchannel2025} enhance global modeling by rearranging input patches based on space-filling curves. 
Despite these advances, existing methods rely on fixed token-ordering schemes that struggle to preserve dynamic spatial proximity and inadequately model dependencies within the spatial domain.


%
%

\section{Method}

In this section, we first formalize the point-level collaborative perception pipeline to establish  our \ourname framework, and then provide a detailed exposition of each submodule’s functionality.

\subsection{Overview}

As illustrated in Fig. \ref{fig:structure}, we consider a multi-vehicle collaborative driving scenario with $N$ agents, indexed by $i\in \mathcal{S}=\{1,\dots,N\}$. For convenience, let $i = \mathrm{e}$ denote the ego agent. The local point cloud captured by agent $i$ is represented as
$
\mathcal{O}_i \in \mathbb{R}^{m\times c},
$
where $m$ is the number of points and $c$ the number of feature channels.
The collaborative perception pipeline then proceeds as follows:


\begin{subequations}
\vspace{-5mm}
\begin{align}
	& \mathcal{X}_i = \mathit{\Phi}_{pt}(\mathcal{O}_i), 
	\label{eq:framework:tokenizer} \\ 
	& \mathcal{F}_i = \mathit{\Phi}_{enc}(\mathcal{X}_i), 
	\label{eq:framework:enc}  \\
	& \mathcal{M}_i = \left[ \mathcal{F}_i, \mathcal{P}_i, \xi_i  \right], 
	\label{eq:framework:share}  \\
	& \mathcal{G}_i = \mathit{\Phi}_{agg}\left(\mathcal{M}_i, \{ \mathcal{M}_j,  \Gamma(\xi_i, \xi_j)  \}  \right),   j \neq i, 
	\label{eq:framework:agg}  \\
	& \mathcal{H}_i = \mathit{\Phi}_{fuse}(\mathcal{G}_i),  \\
	& \mathcal{\hat{Y}}_i = \mathit{\Phi}_{task}\left( \mathcal{H}_i  \right) .  
\end{align}  
\vspace{-5mm}
\label{eq:framework}
\end{subequations}

To transform the raw point-cloud data into a one-dimensional token sequence, we introduce the point tokenizer module (\myeq{eq:framework:tokenizer}). This module addresses the inherent uneven spatial distribution, redundancy, and noise within raw point clouds by placing sampling points at fixed intervals in the 3D space. For each sampling point associated with neighboring raw points within its grid, the point tokenizer captures the semantic composition of multiple surrounding raw points by aggregating features such as spatial position distribution, intensity distribution, and local point density into a single token representation.
This effectively highlights salient information and reduces the initial sequence length, achieving an optimal balance between structural-semantic expressiveness and computational efficiency.
 The resulting output is a point-level token feature sequence $\mathcal{X}_i \in \mathbb{R}^{l\times d}$, where $l$ denotes the number of tokens and $d$ indicates the embedding dimension.

Then, agent $i$ processes its token sequence using a semantic-aware dynamic encoder (\myeq{eq:framework:enc}), applying learned semantic importance scores to retain only the most object-relevant tokens. This results in a compact yet comprehensive feature sequence $\mathcal{F}_i \in \mathbb{R}^{k_i\times d}$, where $k_i$ denotes the number of tokens selected. These selected tokens, together with their spatial coordinates $\mathcal{P}_i$ and the agent's pose information $\xi_i$, are encapsulated into the message packet $\mathcal{M}_i$ for transmission and sharing (\myeq{eq:framework:share}).

Upon receiving message $\mathcal{M}_j$ from a neighboring agent $j$, the ego agent employs point aggregation (\myeq{eq:framework:agg}). Specifically, it first computes the coordinate-transformation matrix using the operator $\Gamma(\cdot)$ based on the relative poses of agents $j$ and $i$. Subsequently, the spatial coordinates of agent $j$'s tokens are transformed into the ego-centric coordinate frame. These transformed tokens are concatenated into a unified sequence $\mathcal{G}_i \in \mathbb{R}^{({\sum_j k_j})\times d}$, prepared for feature fusion and downstream reasoning processes. 

Subsequently, the semantic-aware dynamic fusion module integrates cross-agent tokens to capture large-scale spatial feature information, generating a refined sequence $\mathcal{H}_i \in \mathbb{R}^{(\sum_j k_j)\times d}$, which is then fed into the downstream task network to produce the final prediction $\hat{\mathcal{Y}}_i$.




\subsection{Samentic-aware Dynamic Encoder}

The semantic-aware dynamic encoder module comprises several stacked semantic-guided dynamic state space blocks and concludes with a semantic-importance filtering stage.

 For agent $i$, the primary objective during encoding process is to enhance object-related foreground features and suppress background and noise, facilitating subsequent selection of key tokens for reliable feature communication and fusion. To capture fine-grained semantics at the token level, efficiently process very long token sequences, and model long-range spatial dependencies, we introduce the semantic-guided dynamic state space block (SDSSB), as illustrated in \myfig{fig:structure}(a). This module exhibits dynamic characteristics in two aspects. First, it dynamically reorders tokens based on semantic grouping using the semantic-aware token reordering module (STR) (\myfig{fig:structure}(c)). Second, it extends the conventional state space model by incorporating dynamic frequency-domain representations, as implemented in the frequency-enhanced state space model (FSSM) (\myfig{fig:structure}(d)). 
In addition, we incorporate a scene context extraction submodule (\myfig{fig:structure}(b)) to extract comprehensive scene-level contextual information.
In the following, we describe each of the aforementioned submodules.


\textbf{Scene Context Extraction.}
The scene context extraction module constructs a refined global context by projecting 3D point-level tokens onto a 2D $xy$ grid, applying grid-wise average and max pooling, and refining the resulting feature map through a convolutional layers. This holistic representation directs the semantic-aware token reordering module to focus on token-level features and facilitates semantic analysis at both the scene and token levels. It also supplies the frequency-enhanced state space model with dynamically extracted frequency-domain cues that enrich token representations across spatial and spectral dimensions.



\textbf{Semantic-aware Token Reordering.}
Before applying state space modeling, 3D point-level tokens must be serialized into a one-dimensional sequence while preserving spatial locality. However, conventional reordering schemes such as raster scans or space-filling curves often require costly multi-directional passes and struggle to maintain spatial proximity, particularly in dynamic scenes with sparse foreground objects. These fixed patterns impose strong inductive biases and lack adaptability to diverse scene geometries, frequently placing semantically similar and spatially adjacent tokens far apart in the sequence, thereby limiting effective interaction and representation.

As illustrated in \myfig{fig:structure}(b), we introduce a semantic-aware token reordering module in which each token is augmented with a scene dynamic prompt $G_s$ that integrates global scene-level context into its local semantic representation, enabling each token to determine its semantic-group membership more reliably based on richer contextual evidence.
For interpretability  and stability during training, the prompt is factorized as \myeq{eq:reordering}. The matrix $G_{\mathrm{sp}}$ serves as a scene-specific prompt, derived from the scene-context feature $F_{\mathrm{sc}}$ through convolution and projection. In contrast, $G_{\mathrm{ss}}$ is a fixed, scene-shared prompt that preserves a consistent semantic structure across different scenes.
$r$ is the low-rank dimension , ensuring that $G_s$ remains both expressive and well regularized. 
Subsequently, $G_s$ is average-pooled, reshaped, and added to every token in the input sequence $F$.
\begin{equation}
\begin{aligned}
	& G_s = G_{\mathrm{sp}} \otimes  G_{\mathrm{ss}}, 
	G_{\mathrm{sp}}\in\mathbb{R}^{d_{\mathrm{sp}}\times r}, 
	G_{\mathrm{ss}}\in\mathbb{R}^{r\times d}.
\end{aligned}
\label{eq:reordering}
\end{equation}


We further inject dynamic prompts into each token and refine their semantics using a lightweight semantic projection composed of linear layers. This projects the channel dimension from $d$ to $d_g$, where $d_g$ corresponds to the maximum expected number of object instances. 
A subsequent linear layer followed by a softmax operation predicts the semantic group membership probabilities, from which each token is assigned a semantic group index $I^{g}$. Tokens with similar semantic characteristics—such as the front and rear of the same vehicle or adjacent vehicles—are thus assigned identical or closely related indices.
The token sequence is then reordered to produce $F^{\mathrm{ro}}$.
To provide a consistent semantic bias within each group and to allow downstream tasks to indirectly supervise index assignment, we introduce a set of group prompts $G_{g}$. Each token incorporates its corresponding group prompt based on the assigned index $I^{g}$.


Finally, a semantic-importance head predicts a saliency score $\hat{s}$, and the final token representation is produced by element-wise modulation of the group-augmented features with this score.
As shown in \myeq{eq:loss_s}, the semantic-importance score is supervised by assigning a positive label ($s^{\mathrm{gt}}=1$) to a token whose coordinate $\mathcal{P}^{xyz}$ lies inside any ground-truth 3D bounding box $b \in \mathcal{B}$; otherwise, the label is zero. This supervision not only promotes accurate saliency importance score prediction but also guides $F^{s}$ toward more discriminative semantic representations.
\begin{equation}
\begin{aligned}
	& s^{\text{gt}} =
			\begin{cases}
				1, & \exists\, b \in \mathcal{B}\;\text{s.t.}\;\mathcal{P}^{xyz}\in b,\\
				0, & \text{otherwise},
			\end{cases} \\ 
	&\mathcal{L}^{\mathrm{s}} = \mathcal{L}_{\text{focal}}\bigl(\hat{s}, s^{\text{gt}}\bigr) .  
\end{aligned}
\label{eq:loss_s}
\end{equation}

\textbf{Frequency-enhanced State Space Model.}
Although Mamba efficiently models long‐range dependencies in images and indoor point clouds, its performance degrades in large outdoor LiDAR scenes where sparse, small‐scale vehicle points are easily confused with background clutter. Inspired by psychophysical evidence that biological vision isolates camouflaged objects via frequency‐selective filtering \cite{acklehFrequencydependentEvolutionPredator2021,tanFrequencyawareDeepfakeDetection2024a}, we introduce a compact spectral descriptor that highlights vehicle‐specific contours while suppressing broadband terrain noise. Leveraging the formal similarities between the state-space matrix $C$ and the query matrix of linear attention, we inject these frequency features directly into matrix $C$ (detailed mathematical derivation is provided in \textit{Suppl.} \ref{suppl:fssm}.), preserving Mamba’s $O(N)$ complexity. The resulting frequency-enhanced state space model effectively captures long-range dependencies and spectral correlations, thereby more reliably disentangling object-related foreground tokens from pervasive background clutter.


To extract frequency-domain features, we first preprocess the feature map $F_{\mathrm{sc}}$ produced by the scene-context extraction submodule, which captures the global contextual information of the scene.
 A 2D convolution followed by layer normalization adjusts the channel dimension, yielding $X_{\text{sc}}\in\mathbb{R}^{d \times h \times w}$, where $d$, $h$ and $w$ denote the channel, height and width of the feature map.
We then apply a 2D Discrete Fourier Transform (DFT) to the local window of $X_{\text{sc}}$, capturing the spatial frequency content in the vicinity of each token’s location, as follows:
\begin{equation}
\begin{aligned}
X&_{sc} = \operatorname{Norm}\bigl(\operatorname{Conv2D}(F_{sc})\bigr), \\
F&^{\mathrm{win}}_{x_j, y_j} = \operatorname{Win}\bigl( X_{sc}, (x_j, y_j), (H_k, W_k) \bigr), \\
Q&_{x_j, y_j} = \operatorname{Freq}\bigl(  \operatorname{DFT}(F^{\text{win}}_{x_j, y_j})\bigr),   \\
Q&^{freq} = \bigl[ Q_{x_j, y_j}, ... \bigr],
\end{aligned}
\end{equation}
here, $j=1,\dots,M$, with $M$ denoting the length of the token sequence. The operator $\mathrm{Win}(\cdot)$ extracts a window patch of size $(H_k,W_k)$ from the 2D feature map centered at $(x_j,y_j)$, and $\mathrm{Freq}(\cdot)$ extracts both low- and high-frequency components from this window and aggregates them into a unified frequency-domain feature.


Consequently, we obtain the following  equation of frequency-enhanced state space model:
\begin{equation}
\begin{aligned}
  h_i &= \overline{A}_i\,h_{i-1} + \overline{B}_i\bigl(\Delta x_i\bigr), \\
  y_i &= \Bigl(C_i + \gamma\,Q^{freq}_i\Bigr)\,h_i + D\,x_i,
\end{aligned}
\label{eq:fssme}
\end{equation}
where $\gamma>0$ is a learnable scaling factor that balances the numerical range. $x_i$ denotes the input token in sequence form, $y_i$ denotes the output token, and all other symbols represent parameters of the SSM.

Furthermore, we adopt a dual-scope modeling strategy  \cite{jinUniMambaUnifiedSpatialchannel2025,zhangVoxelMambaGroupfree2024}. In the global scope, the FSSM processes the entire token sequence to capture long-range spatial dependencies. In the local scope, the FSSM operates on smaller windows to model fine-grained structures. Combining both scopes yields richer and more comprehensive feature representations.

\textbf{Semantic-importance filtering.}
The semantic-importance filtering module is the final stage of the encoding pipeline. It ranks tokens by semantic significance and selects the top $k$ tokens that carry the richest foreground object information. This ensures that only the most relevant tokens are shared, enabling reliable collaboration under severe bandwidth constraints. Since the sequence has already undergone comprehensive semantic analysis, filtering requires only a few linear layers and a sigmoid activation to compute the semantic-importance scores.


\subsection{Semantic-aware Dynamic Fusion}
%
%
In noisy real-world environments, localization errors degrade the accuracy of coordinate transformations, causing neighbor-agent tokens to misalign in the ego-centric coordinate and distort feature representations. 
 Our fusion pipeline’s SDSSB blocks can perceive and partially compensate for this by dynamically analyzing semantic similarity at both scene and token levels and by modeling long-range dependencies in spatial and spectral domains. To further enhance our method’s adaptive misalignment correction, we introduce a neighbor-to-ego alignment module that dynamically adjusts token positions based on local spatial consistency and explicitly realigns spatially displaced point-level tokens.

\vspace{-2mm}
\begin{figure}[!htb]
    \centering
    \includegraphics[width=0.42\textwidth]{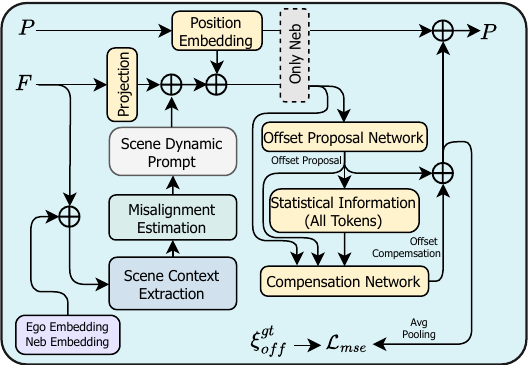}
    \caption{The  architecture  of Neighbor-to-Ego Alignment Module.
}
    \label{fig:align}
\end{figure} 
\vspace{-3mm}

\textbf{Neighbor-to-ego alignment.}
This module implements a closed-loop alignment process that transitions from agent-level coordination to token-level fine-tuning, as shown in \myfig{fig:align}. 
First, we integrate ego and neighbor embeddings into the mixed token sequence to distinguish each agent’s origin. Next, the scene-context extraction module $\Phi_{\mathrm{sc}}$ produces both a fused global scene feature map $F_{\mathrm{sc}}^{\mathrm{fuse}}$ and individual per-agent maps $F_{\mathrm{sc}}^{\mathrm{agent}}$. 
We then estimate each neighbor’s overall spatial displacement relative to the ego agent using $\Phi_{\mathrm{mis}}(\cdot)$. We convert this displacement into a scene-dynamic prompt $G_s$ via $\Phi_{\mathrm{prompt}}(\cdot)$ and inject it into the neighbor’s tokens.
\begin{equation}
\begin{aligned}
  & F_{sc}^{\mathrm{fuse}}, F_{sc}^{\mathrm{agent}} = \Phi_{\mathrm{sc}}\bigl( F \bigr), \\
  & F_{sc}^{\mathrm{agent}} = \bigl[F_{sc}^{(i)}, ... \bigr], \quad i=1, ...,N, \\
  & G_s^{(j)} = \Phi_{\mathrm{prompt}}\bigl(  \Phi_{\mathrm{mis}}\bigl(  
  F_{sc}^{\mathrm{fuse}}, F_{sc}^{(\mathrm{e})} , F_{sc}^{(j)} \bigr) \bigr), j \neq \mathrm{e}  .
\end{aligned}
\label{eq:align1}
\end{equation}


Next, each token generates an offset proposal $\Delta_{p}$ according to both the agent-level misalignment and its own semantic features. We aggregate all token-level proposals for each agent to compute a statistical offset bias at the agent level, characterized by the mean and standard deviation. Finally, each token adjusts its initial proposal toward this global bias to produce a compensated offset $\delta_{p}$. We update the original token coordinates $P$ by adding the offset proposal $\Delta_{p}$ and the compensated offset $\delta_{p}$, resulting in the adjusted coordinates $P + \left( \Delta_{p} + \delta_{p} \right)$.


\begin{equation}
\begin{aligned}
  & \Delta_{p} = \Phi_{\mathrm{proposal}}\bigl( F^p \bigr), \\
  & \mu, s = \Phi_{\mathrm{statistic}}\bigl( \Delta_{p} \bigr), \\
  & \delta_{p} = \Phi_{\mathrm{compensate}}\bigl(  F^p, \Delta_{p}, \mu, s \bigr), \\
  & P^{out} = P + \bigl( \Delta_{p} + \delta_{p} \bigr).
\end{aligned}
\label{eq:align2}
\end{equation}


To supervise the alignment process, we compute a reference ground-truth offset $\xi_{\text{off}}^{\mathrm{gt}}$ by comparing the true and noisy spatial coordinates of the ego and neighbor agents. The resulting alignment loss is defined as:
\begin{equation}
\begin{aligned}
\mathcal{L}^{\mathrm{off}} = \mathcal{L}_\mathrm{mse}\left(  \left(\Delta_{p} + \delta_{p}  \right)  ,\,\xi_{\text{off}}^{\mathrm{gt}}\right).
\end{aligned}
\end{equation}


\section{Experiment}

\begin{table}[]
  \centering
  \caption{Comparison of 3D detection performance on three datasets.}
  \label{table:ap-comparison}
  \small
  \begin{tabular}{@{}c  c  c  c@{}}
    \toprule
    \multirow{2}{*}{Method}
      & \multicolumn{1}{c}{OPV2V}
      & \multicolumn{1}{c}{V2V4Real}
      & \multicolumn{1}{c}{DAIR-V2X} \\ 
    & {AP@0.5/0.7} & {AP@0.5/0.7} & {AP@0.5/0.7} \\
    \midrule
    AttFuse         & 0.924 / 0.793    & 0.544 / 0.309    & 0.548 / 0.371    \\
    V2VNet          & 0.930 / 0.821    & 0.402 / 0.256    & 0.635 / 0.392    \\
    V2X-ViT         & 0.963 / 0.883    & 0.540 / 0.311    & 0.698 / 0.446    \\
    CoBEVT          & 0.957 / 0.838    & 0.556 / 0.394    & 0.623 / 0.382    \\
    Where2comm      & 0.947 / 0.846    & 0.512 / 0.324    & 0.616 / 0.401    \\
    SiCP            & 0.949 / 0.863    & 0.488 / 0.310    & 0.675 / 0.417    \\
    CoSDH           & 0.949 / 0.869    & 0.524 / 0.337    & 0.702 / 0.484    \\
    CollaMamba      & 0.947 / 0.892    &  0.582 / 0.372   & 0.662 / 0.442    \\
    CoMamba         & 0.933 / 0.842    &  0.537 / 0.361   & 0.602 / 0.495    \\
    \rowcolor[HTML]{EFEFEF}
    \ourname           & 0.973 / 0.934    & 0.644 / 0.447    & 0.761 / 0.593    \\
    \bottomrule
  \end{tabular}
\end{table}


\subsection{Settings}
To validate the effectiveness of the proposed method, we conducted extensive experiments on several simulated and real-world datasets, including OPV2V \cite{xuOPV2VOpenBenchmark2022a}, V2V4Real \cite{xuV2V4RealRealWorldLargeScale2023a}, and DAIR-V2X \cite{yuDAIRV2XLargeScaleDataset2022}. We compared our approach with recent state-of-the-art methods, such as AttFuse \cite{xuOPV2VOpenBenchmark2022a}, V2VNet \cite{wangV2VNetVehicletoVehicleCommunication2020}, V2X-ViT \cite{xuV2XViTVehicletoeverythingCooperative2022}, CoBEVT \cite{xuCoBEVTCooperativeBirds2022}, Where2comm \cite{huWhere2commCommunicationEfficientCollaborative2022b}, SiCP \cite{quSiCPSimultaneousIndividual2024}, CoSDH \cite{xuCoSDHCommunicationefficientCollaborative2025}, CollaMamba \cite{liCollaMambaEfficientCollaborative2024}, and CoMamba \cite{liCoMambaRealtimeCooperative2024}.
We evaluated collaborative 3D object detection performance using average precision (AP) at intersection over union (IoU) thresholds of 0.5 and 0.7. Communication volume was measured in bytes and expressed as a base-2 logarithm, with each floating-point number represented by 4 bytes during transmission. Additionally, ``FLOPs" was used to assess the computational complexity of the model.

%

\begin{figure*}[t]
    \centering
    \begin{subfigure}[t]{0.33\textwidth}
        \centering
        \includegraphics[width=\linewidth]{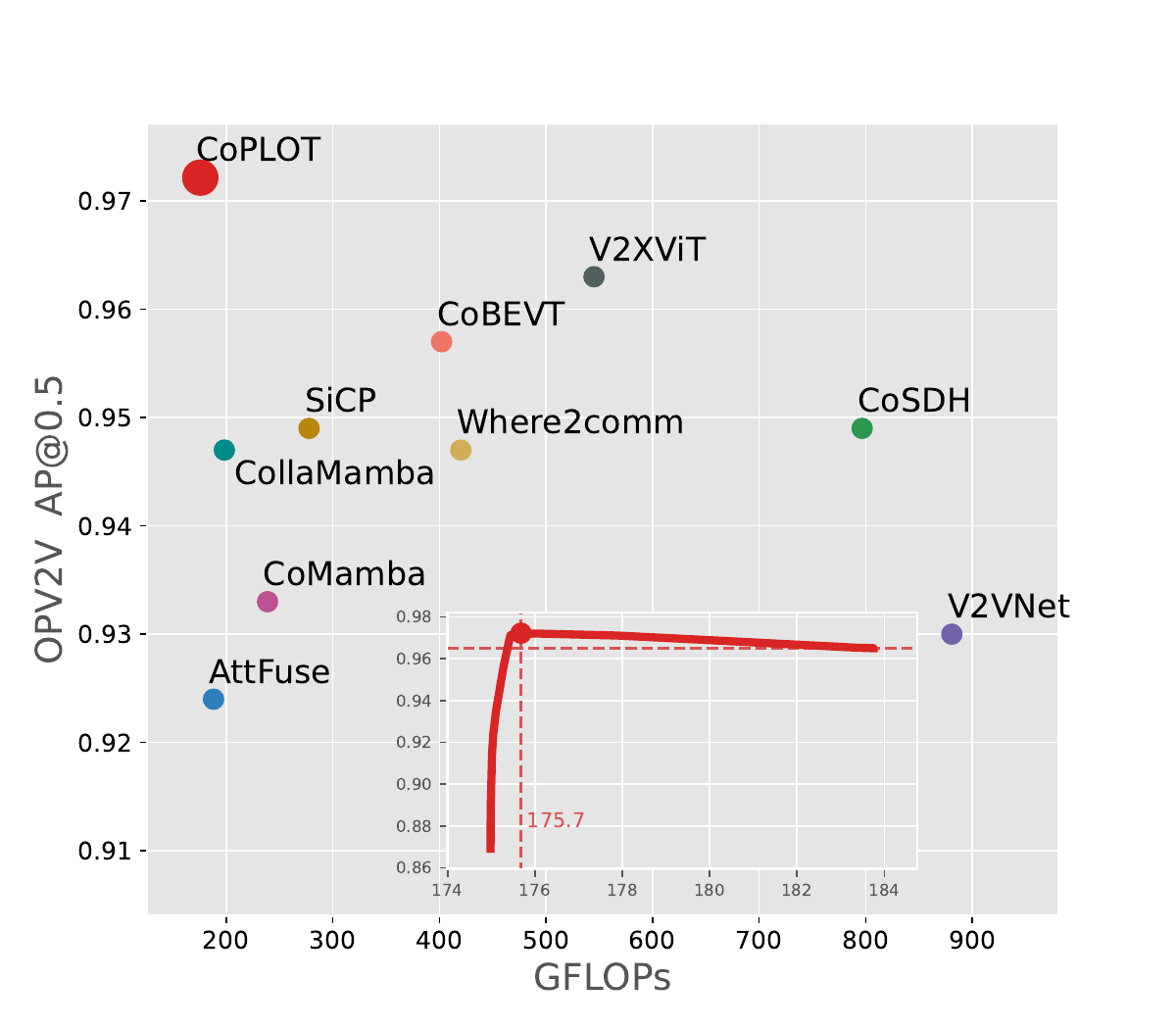}
    \end{subfigure}
    \begin{subfigure}[t]{0.33\textwidth}
        \centering
        \includegraphics[width=\linewidth]{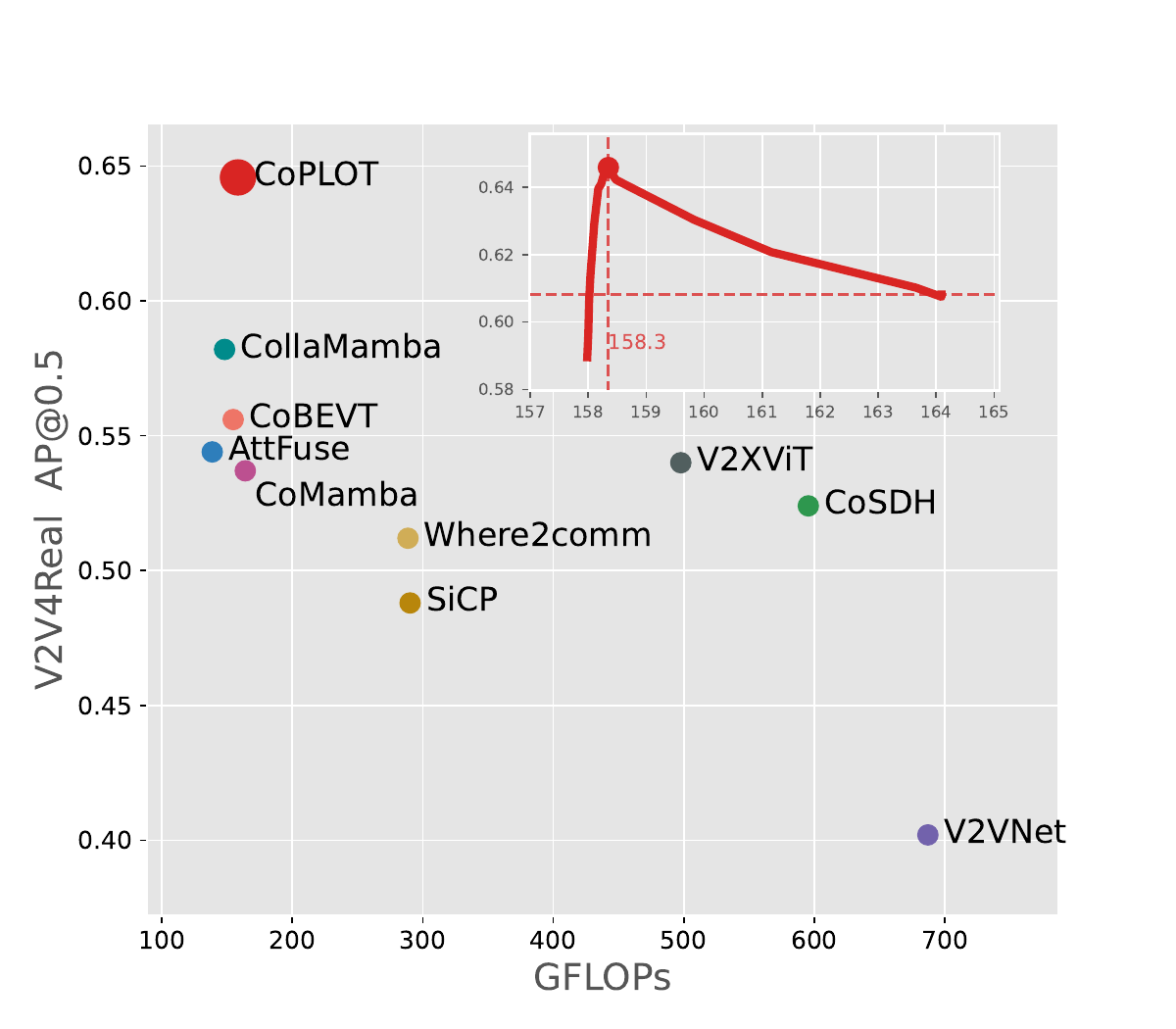}
    \end{subfigure}
    \begin{subfigure}[t]{0.33\textwidth}
        \centering
        \includegraphics[width=\linewidth]{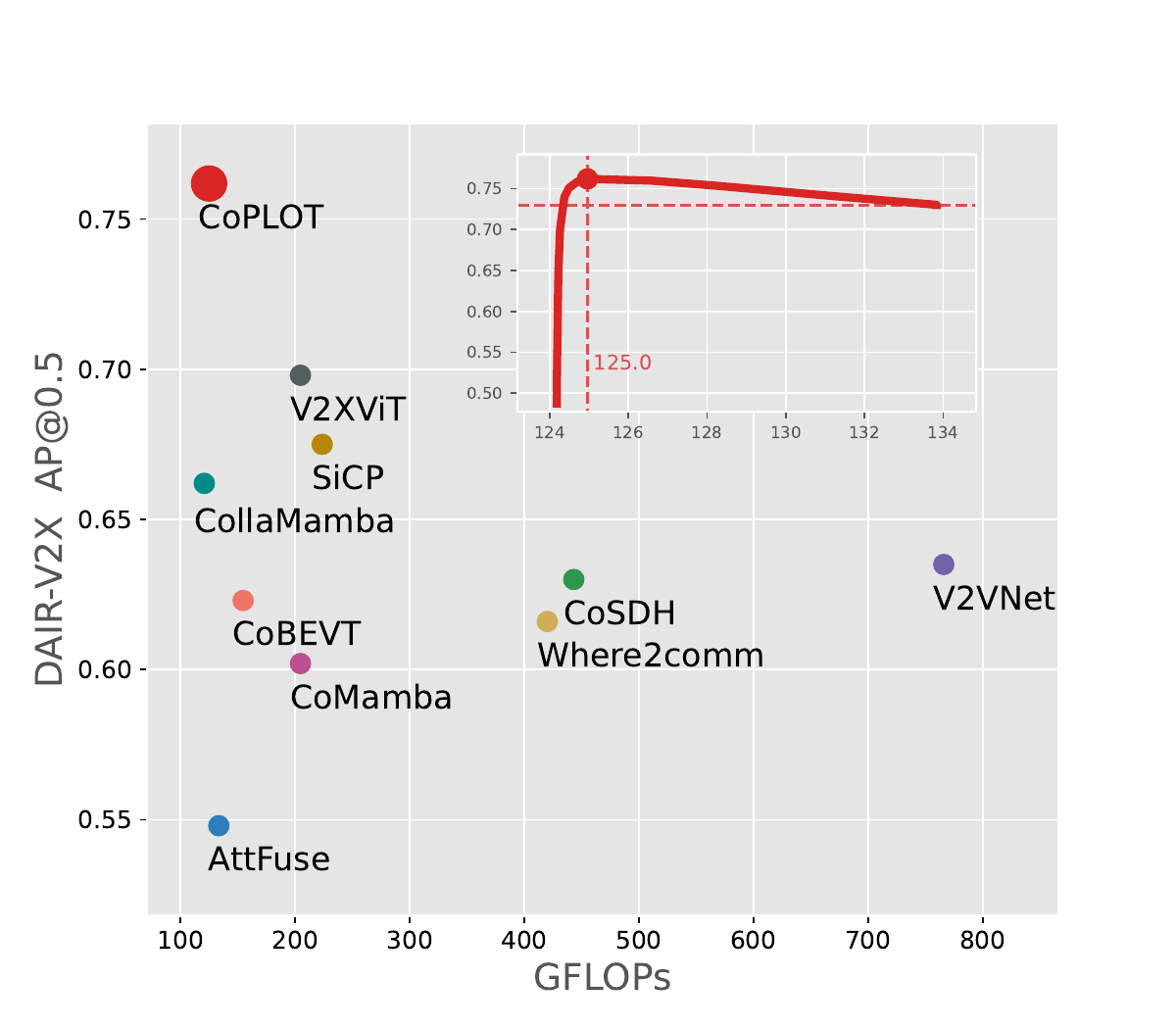}
    \end{subfigure}
    \caption{Our method achieves a favorable balance between perception performance and computational overhead  (measured in GFLOPs) across all three collaborative perception datasets. The red curve is obtained by varying the number of point-level tokens used for computation to correspond to different computational budgets.
    }
    \label{fig:compt50}

\end{figure*}

\begin{figure*}[!htb]
  \centering
  \subfloat[Computational overhead vs. neb top-$k$ \label{fig:computation_all_compt_flops_ntoken}]{
    \includegraphics[width=0.32\textwidth]{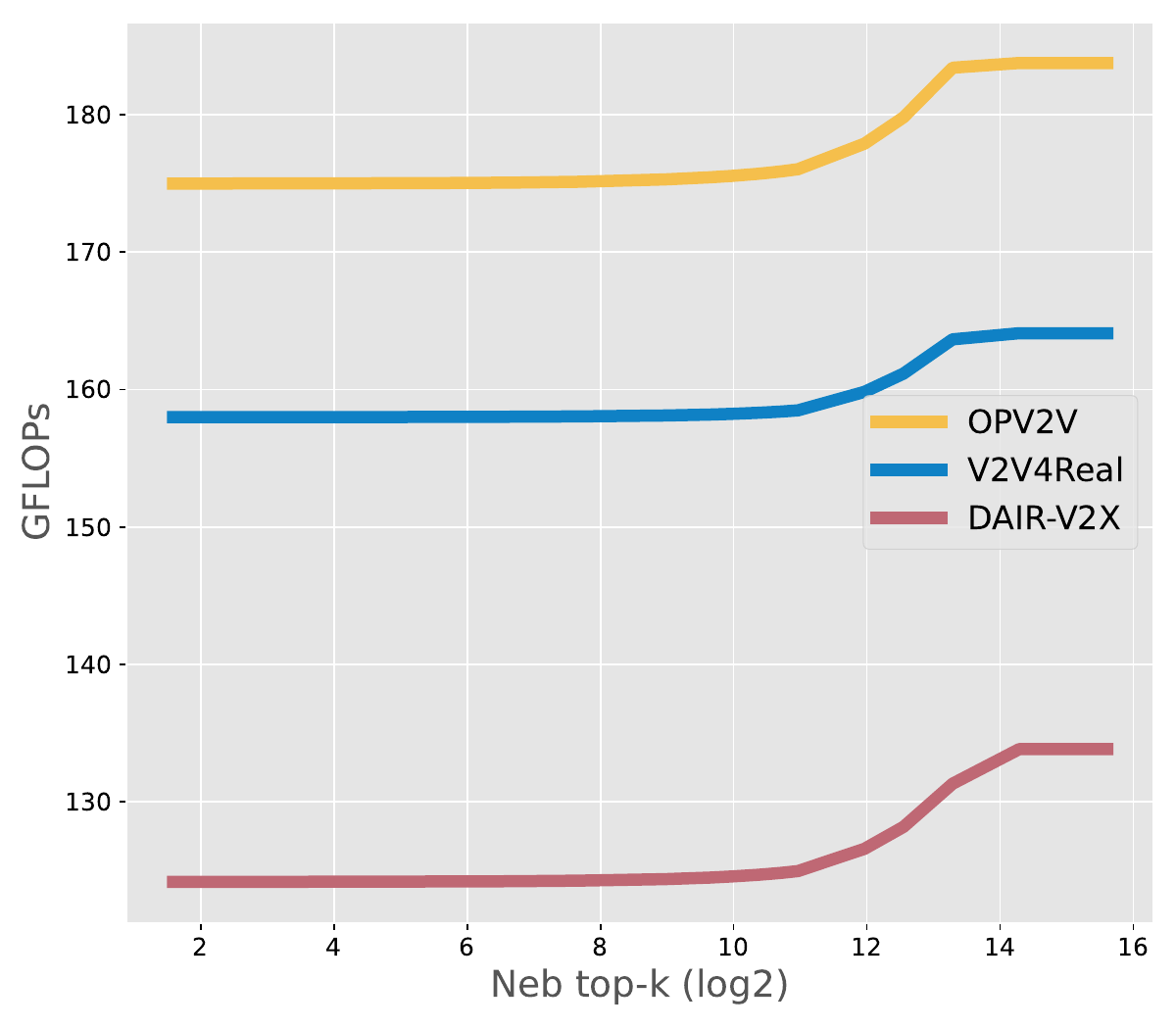}
  }\hfill
  \subfloat[Performance vs. communication overhead  \label{fig:communication_dairv2x_ap_50}]{
    \includegraphics[width=0.32\textwidth]{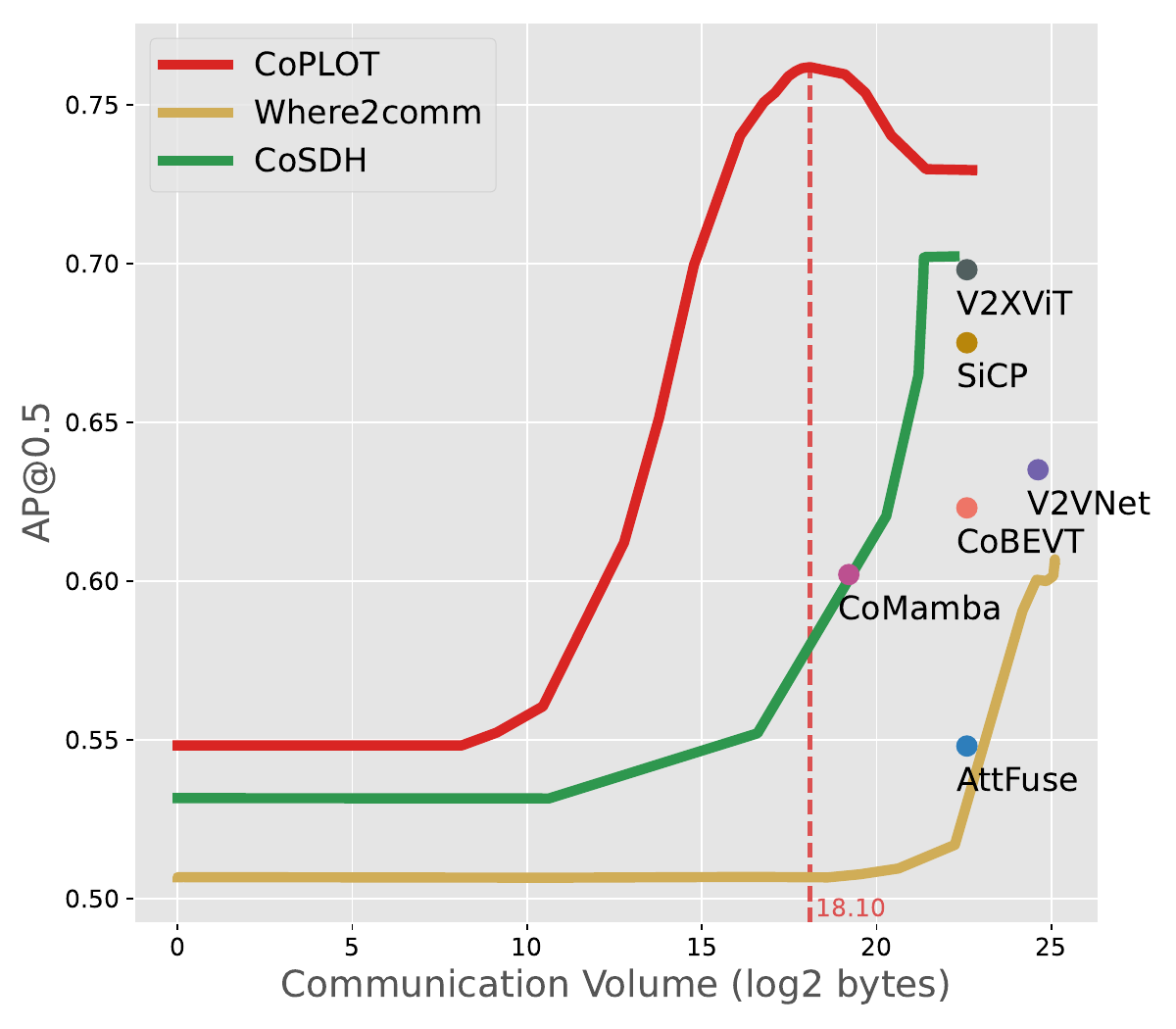}
  }\hfill
  \subfloat[Robustness to localization noise \label{fig:location_v2v4real_ap_50}]{
    \includegraphics[width=0.32\textwidth]{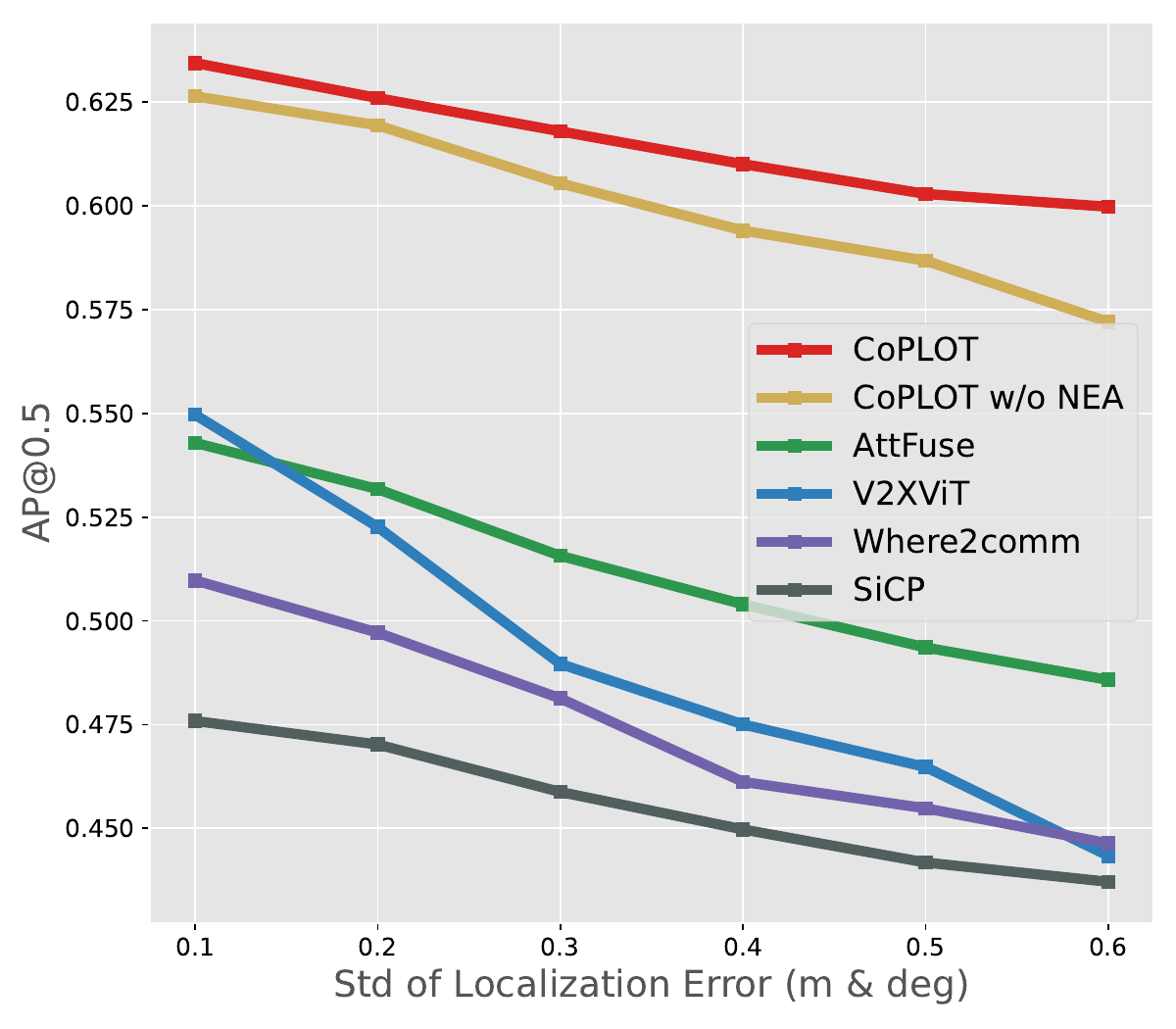}
  }
  \caption{%
  More comparisons on the three datasets: 
    (a) Relationship between the number of top-$k$ neighbors selected and communication overhead on the three datasets,  
    (b) Comparison of the trade-off between perception performance and communication overhead on DAIR-V2X,  
    (c) Robustness to localization noise on V2V4Real.
  }
  \vspace{-6mm}
\end{figure*}

\subsection{Comparison Analysis}
\textbf{Perception Performance.} 
As shown in \mytab{table:ap-comparison}, we compare our method with several baseline approaches and conduct extensive experiments across three datasets, following the experimental setups of existing methods \cite{huWhere2commCommunicationEfficientCollaborative2022b,liCollaMambaEfficientCollaborative2024}. The results demonstrate that our \ourname outperforms state-of-the-art methods on all three datasets in the 3D detection task. 
Specifically, the performance surpasses the second-best result by 4.2\%, 5.3\%, and 9.8\% in terms of AP@0.7 on OPV2V, V2V4Real, and DAIR-V2X, respectively.
 Notably, significant improvements were observed on two challenging real-world datasets, indicating that our method, which processes point-level tokens containing comprehensive structural and semantic information, provides more effective cues for distinguishing objects in complex scenarios. 
This validates the effectiveness of the proposed framework, which leverages point-level optimized tokens for collaborative perception, in both simulated and real-world environments.

\textbf{Computation Overhead.} 
\myfig{fig:compt50} illustrates the trade-off between perception performance and computational overhead. Our \ourname framework enables adaptive selection of point-level tokens for fusion based on a given computational budget, yielding accuracy curves under varying levels of overhead. From the figure, we observe the following:
1) Compared to existing methods, \ourname achieves up to 80\% reduction in computational overhead while consistently maintaining high perception accuracy on the three datasets.
 The high computation efficiency is achieved by selectively involving a compact set of potential object-related tokens in computation.
Moreover, the point-level tokens retain comprehensive structural details and rich semantic information, enabling the compact token sequence to achieve strong performance with minimal computational cost.
2) The framework is particularly well-suited for deployment in real-world scenarios with dynamically constrained computational resources. It can flexibly adjust the number of critical tokens used for computation according to the available resources.
3) Incorporating too many tokens may not improve perception performance, as excessive background or irrelevant noise introduces unnecessary burden to the model. Therefore, allocating excessive computational resources is unnecessary. 
Furthermore, \myfig{fig:computation_all_compt_flops_ntoken} illustrates the relationship between the number of top-$k$ tokens selected from neighboring agents and the computational overhead. The curve in the figure appears exponential because the x-axis uses a log$_2$ scale for the top-$k$ token counts, indicating a linear relationship between the computational overhead and the number of top-$k$ tokens.


\begin{table}[]
  \centering
  \caption{Ablation study of different model components.}
  \label{table:ablation}
  \small
  \begin{tabular}{@{}lcc@{}}
    \toprule
    \multirow{2}{*}{Configuration}
      & \multicolumn{1}{c}{OPV2V}
      & \multicolumn{1}{c}{V2V4Real}      \\
      & AP@0.5/0.7   & AP@0.5/0.7                      \\
    \midrule
  \rowcolor[HTML]{EFEFEF}
    \ourname                              & 0.973 / 0.934 & 0.644 / 0.447      \\
    -w/o scene specific prompt            & 0.957 / 0.916 & 0.576 / 0.394      \\
    -w/o scene shared prompt              & 0.971 / 0.927 & 0.594 / 0.389      \\
    -w/o group prompt                     & 0.961 / 0.925 & 0.565 / 0.355      \\
    -w/o dynamic prompt                   & 0.952 / 0.910 & 0.510 / 0.349      \\
    -w/o frequency enhancement            & 0.963 / 0.899 & 0.492 / 0.321      \\
    \midrule
    -w/o STR                              & 0.942 / 0.871 & 0.578 / 0.409      \\
    -w/o FSSM                             & 0.954 / 0.889 & 0.521 / 0.331      \\
    -w/o SCE                              & 0.950 / 0.909 & 0.538 / 0.324      \\
    \bottomrule
  \end{tabular}
\end{table}

\textbf{Communication Overhead.} 
\myfig{fig:communication_dairv2x_ap_50} presents the performance comparison under varying communication budgets. \ourname achieves a superior trade-off between perception performance and communication overhead compared to baseline methods by transmitting a more compact and informative token sequence. 
During encoding, \ourname incorporates semantic information from both the global scene and individual tokens while capturing long-range dependencies in spatial and spectral domains, enabling more effective selection of critical object-related tokens for transmission.

\textbf{Robustness to Localization Noise.} 
In real-world scenarios, localization noise can cause misalignment in collaborative perception. To evaluate robustness under such conditions, we conduct experiments by adding Gaussian noise to the agents' poses, as illustrated in \myfig{fig:location_v2v4real_ap_50}. The results show that \ourname consistently maintains state-of-the-art performance across varying levels of noise. Additionally, we perform an ablation study by removing the neighbor-to-ego alignment (NEA) module, which results in a noticeable drop in robustness. This validates the effectiveness of the NEA module in mitigating localization-induced misalignment.

\subsection{Ablation Study}


We further conducted ablation studies to evaluate the contribution of each component in the \ourname framework. As shown in \mytab{table:ablation}, the upper half of the table reports results when individual submodules are removed, while the lower half examines the impact of excluding larger structural components. The results demonstrate that removing any single submodule leads to a performance drop, indicating that all components contribute to the overall effectiveness of the model. 
Notably, removing the semantic-aware token reordering module, the frequency enhancement module, or the scene-specific prompt leads to a marked performance drop. This highlights the critical role of semantic analysis that integrates global scene context with local token features, and the necessity of jointly modeling long-range dependencies across spatial and spectral domains. 
These components enhance the model’s ability to differentiate object-related semantic attributes, enriching the point-level tokens and ultimately improving perception performance.
Additional ablation experiments and detailed analyses are available in the Supplementary Material.

\section{Conclusion}
In this paper, we propose \ourname, a novel collaborative perception framework that leverages point-level optimized tokens as intermediate representations to address the limitations of conventional BEV features, such as the loss of structural information and high computational and communication costs. We develop a tailored point-native processing pipeline along with key modules, including a semantic-aware token reordering module, a frequency-enhanced state space model, and a neighbor-to-ego alignment module. These components are specifically designed to support the generation, transmission, and fusion of point-level token sequences. Extensive experiments validate the superior performance of \ourname. 
Future work includes extending our framework to more complex 3D collaborative perception tasks, as well as exploring joint optimization of perception accuracy with both communication and computational resource constraints.

%
%

\bibliography{aaai2026}

\begin{thebibliography}{55}
\providecommand{\natexlab}[1]{#1}

\bibitem[{Ackleh and Veprauskas(2021)}]{acklehFrequencydependentEvolutionPredator2021}
Ackleh, A.~S.; and Veprauskas, A. 2021.
\newblock Frequency-Dependent Evolution in a Predator--Prey System.
\newblock \emph{Natural Resource Modeling}, 34(3): e12308.

\bibitem[{Ali, Zimerman, and Wolf(2024)}]{aliHiddenAttentionMamba2024}
Ali, A.; Zimerman, I.; and Wolf, L. 2024.
\newblock The Hidden Attention of Mamba Models.
\newblock arXiv:2403.01590.

\bibitem[{Bai et~al.(2024)Bai, Wu, Barth, Liu, Akin~Sisbot, Oguchi, and Huang}]{baiSurveyFrameworkCooperative2024}
Bai, Z.; Wu, G.; Barth, M.~J.; Liu, Y.; Akin~Sisbot, E.; Oguchi, K.; and Huang, Z. 2024.
\newblock A {{Survey}} and {{Framework}} of {{Cooperative Perception}}: {{From Heterogeneous Singleton}} to {{Hierarchical Cooperation}}.
\newblock \emph{IEEE Transactions on Intelligent Transportation Systems}, 25(11): 15191--15209.

\bibitem[{Baumbach(2010)}]{baumbachPsychophysicsHumanVision2010}
Baumbach, J. 2010.
\newblock Psychophysics of Human Vision: The Key to Improved Camouflage Pattern Design.

\bibitem[{Chen et~al.(2024)Chen, Wu, Chitta, Jaeger, Geiger, and Li}]{chenEndtoendAutonomousDriving2024}
Chen, L.; Wu, P.; Chitta, K.; Jaeger, B.; Geiger, A.; and Li, H. 2024.
\newblock End-to-End Autonomous Driving: Challenges and Frontiers.
\newblock \emph{IEEE Trans. Pattern Anal. Mach. Intell.}, 46(12): 10164--10183.

\bibitem[{Das and Geisler(2022)}]{dasCamouflageDetectionExperiments2022}
Das, A.; and Geisler, W. 2022.
\newblock Camouflage Detection: Experiments and a Principled Theory.
\newblock \emph{Journal of Vision}, 22(14): 4069.

\bibitem[{Das and Geisler(2023)}]{dasPredictingHumanCamouflage2023}
Das, A.; and Geisler, W.~S. 2023.
\newblock Predicting Human Camouflage Detection with a Principled Computational Model.
\newblock \emph{Journal of Vision}, 23(9): 5530.

\bibitem[{Dosovitskiy et~al.(2017)Dosovitskiy, Ros, Codevilla, Lopez, and Koltun}]{dosovitskiyCARLAOpenUrban2017}
Dosovitskiy, A.; Ros, G.; Codevilla, F.; Lopez, A.; and Koltun, V. 2017.
\newblock {{CARLA}}: An Open Urban Driving Simulator.
\newblock In \emph{Proceedings of the 1st {{Annual Conference}} on {{Robot Learning}}}, volume~78 of \emph{Proceedings of {{Machine Learning Research}}}, 1--16.

\bibitem[{Gao et~al.(2024)Gao, Zhang, Lu, Huang, Yang, Xiong, and Liu}]{gaoSurveyCollaborativePerception2024}
Gao, X.; Zhang, X.; Lu, Y.; Huang, Y.; Yang, L.; Xiong, Y.; and Liu, P. 2024.
\newblock A Survey of Collaborative Perception in Intelligent Vehicles at Intersections.
\newblock \emph{IEEE Transactions on Intelligent Vehicles}, 1--20.

\bibitem[{Gu et~al.(2025)Gu, Meng, Zheng, Sun, Ji, Ruan, Cao, and Ji}]{guEfficientMixedHeterogeneous2025}
Gu, Y.; Meng, Y.; Zheng, K.; Sun, X.; Ji, J.; Ruan, W.; Cao, L.; and Ji, R. 2025.
\newblock An Efficient and Mixed Heterogeneous Model for Image Restoration.
\newblock arXiv:2504.10967.

\bibitem[{Han et~al.(2024{\natexlab{a}})Han, Wang, Xia, Han, Pu, Ge, Song, Song, Zheng, and Huang}]{hanDemystifyMambaVision2024}
Han, D.; Wang, Z.; Xia, Z.; Han, Y.; Pu, Y.; Ge, C.; Song, J.; Song, S.; Zheng, B.; and Huang, G. 2024{\natexlab{a}}.
\newblock Demystify Mamba in Vision: A Linear Attention Perspective.
\newblock In \emph{Advances in {{Neural Information Processing Systems}} 38: {{Annual Conference}} on {{Neural Information Processing Systems}} 2024, {{Neurips}} 2024, {{Vancouver}}, {{BC}}, {{Canada}}, {{December}} 10 - 15, 2024}.

\bibitem[{Han et~al.(2024{\natexlab{b}})Han, Tang, Wang, and Li}]{hanMamba3DEnhancingLocal2024}
Han, X.; Tang, Y.; Wang, Z.; and Li, X. 2024{\natexlab{b}}.
\newblock {{Mamba3D}}: Enhancing Local Features for {{3D}} Point Cloud Analysis via State Space Model.
\newblock In \emph{Proceedings of the 32nd {{ACM International Conference}} on {{Multimedia}}, {{MM}} 2024, {{Melbourne}}, {{VIC}}, {{Australia}}, 28 {{October}} 2024 - 1 {{November}} 2024}, 4995--5004.

\bibitem[{Hu et~al.(2022)Hu, Fang, Lei, Zhong, and Chen}]{huWhere2commCommunicationEfficientCollaborative2022b}
Hu, Y.; Fang, S.; Lei, Z.; Zhong, Y.; and Chen, S. 2022.
\newblock Where2comm: {{Communication-Efficient Collaborative Perception}} via {{Spatial Confidence Maps}}.
\newblock \emph{Advances in Neural Information Processing Systems}, 35: 4874--4886.

\bibitem[{Jin et~al.(2025)Jin, Su, Liu, Ma, Wu, Hui, and Yan}]{jinUniMambaUnifiedSpatialchannel2025}
Jin, X.; Su, H.; Liu, K.; Ma, C.; Wu, W.; Hui, F.; and Yan, J. 2025.
\newblock {{UniMamba}}: Unified Spatial-Channel Representation Learning with Group-Efficient Mamba for {{LiDAR-based 3D}} Object Detection.
\newblock In \emph{{{IEEE}}/{{CVF Conference}} on {{Computer Vision}} and {{Pattern Recognition}}, {{CVPR}} 2025, {{Nashville}}, {{TN}}, {{USA}}, {{June}} 11-15, 2025}, 1407--1417.

\bibitem[{Kingma and Ba(2015)}]{kingmaAdamMethodStochastic2015}
Kingma, D.~P.; and Ba, J. 2015.
\newblock Adam: A Method for Stochastic Optimization.
\newblock In \emph{3rd {{International Conference}} on {{Learning Representations}}, {{ICLR}} 2015, {{San Diego}}, {{CA}}, {{USA}}, {{May}} 7-9, 2015, {{Conference Track Proceedings}}}.

\bibitem[{Lang et~al.(2019)Lang, Vora, Caesar, Zhou, Yang, and Beijbom}]{langPointPillarsFastEncoders2019}
Lang, A.~H.; Vora, S.; Caesar, H.; Zhou, L.; Yang, J.; and Beijbom, O. 2019.
\newblock {{PointPillars}}: Fast Encoders for Object Detection from Point Clouds.
\newblock In \emph{2019 {{IEEE}}/{{CVF Conference}} on {{Computer Vision}} and {{Pattern Recognition}} ({{CVPR}})}, 12689--12697.

\bibitem[{Li et~al.(2025{\natexlab{a}})Li, Hou, Xing, Ma, Sun, and Zhang}]{liOccMambaSemanticOccupancy2025}
Li, H.; Hou, Y.; Xing, X.; Ma, Y.; Sun, X.; and Zhang, Y. 2025{\natexlab{a}}.
\newblock {{OccMamba}}: Semantic Occupancy Prediction with State Space Models.
\newblock In \emph{{{IEEE}}/{{CVF Conference}} on {{Computer Vision}} and {{Pattern Recognition}}, {{CVPR}} 2025, {{Nashville}}, {{TN}}, {{USA}}, {{June}} 11-15, 2025}, 11949--11959.

\bibitem[{Li et~al.(2024{\natexlab{a}})Li, Liu, Li, Xu, Li, Yu, and Tu}]{liCoMambaRealtimeCooperative2024}
Li, J.; Liu, X.; Li, B.; Xu, R.; Li, J.; Yu, H.; and Tu, Z. 2024{\natexlab{a}}.
\newblock {{CoMamba}}: {{Real-time Cooperative Perception Unlocked}} with {{State Space Models}}.
\newblock In \emph{{{arXiv}}.Org}.

\bibitem[{Li et~al.(2025{\natexlab{b}})Li, Li, Wang, He, Wang, Wang, and Qiao}]{liVideoMambaStateSpace2025}
Li, K.; Li, X.; Wang, Y.; He, Y.; Wang, Y.; Wang, L.; and Qiao, Y. 2025{\natexlab{b}}.
\newblock {{VideoMamba}}: State Space Model for Efficient Video Understanding.
\newblock In \emph{European {{Conference}} on {{Computer Vision}}, {{ECCV}} 2024}, 237--255.

\bibitem[{Li et~al.(2021)Li, Ren, Wu, Chen, Feng, and Zhang}]{liLearningDistilledCollaboration2021a}
Li, Y.; Ren, S.; Wu, P.; Chen, S.; Feng, C.; and Zhang, W. 2021.
\newblock Learning {{Distilled Collaboration Graph}} for {{Multi-Agent Perception}}.
\newblock In \emph{Advances in {{Neural Information Processing Systems}}}, volume~34, 29541--29552.

\bibitem[{Li et~al.(2024{\natexlab{b}})Li, Yuan, Luo, Fu, Zhu, Yang, Pan, and Li}]{liCollaMambaEfficientCollaborative2024}
Li, Y.; Yuan, Q.; Luo, G.; Fu, X.; Zhu, X.; Yang, Y.; Pan, R.; and Li, J. 2024{\natexlab{b}}.
\newblock {{CollaMamba}}: {{Efficient Collaborative Perception}} with {{Cross-Agent Spatial-Temporal State Space Model}}.
\newblock arXiv:2409.07714.

\bibitem[{Liang et~al.(2024)Liang, Zhou, Xu, Zhu, Zou, Ye, Tan, and Bai}]{liangPointMambaSimpleState2024}
Liang, D.; Zhou, X.; Xu, W.; Zhu, X.; Zou, Z.; Ye, X.; Tan, X.; and Bai, X. 2024.
\newblock {{PointMamba}}: A Simple State Space Model for Point Cloud Analysis.
\newblock In \emph{Advances in {{Neural Information Processing Systems}} 38: {{Annual Conference}} on {{Neural Information Processing Systems}} 2024, {{Neurips}} 2024, {{Vancouver}}, {{BC}}, {{Canada}}, {{December}} 10 - 15, 2024}.

\bibitem[{Liu et~al.(2023)Liu, Gao, Chen, Peng, Kong, Wang, Xu, Jiang, Xiang, Ma, and Wang}]{liuVehicletoeverythingAutonomousDriving2023}
Liu, S.; Gao, C.; Chen, Y.; Peng, X.; Kong, X.; Wang, K.; Xu, R.; Jiang, W.; Xiang, H.; Ma, J.; and Wang, M. 2023.
\newblock Towards Vehicle-to-Everything Autonomous Driving: A Survey on Collaborative Perception.
\newblock arXiv:2308.16714.

\bibitem[{Liu et~al.(2024{\natexlab{a}})Liu, Huang, Li, Chen, Zhao, Zhao, Zhu, and Zhang}]{liuSelect2ColLeveragingSpatialtemporal2024}
Liu, Y.; Huang, Q.; Li, R.; Chen, X.; Zhao, Z.; Zhao, S.; Zhu, Y.; and Zhang, H. 2024{\natexlab{a}}.
\newblock {{Select2Col}}: Leveraging Spatial-Temporal Importance of Semantic Information for Efficient Collaborative Perception.
\newblock \emph{IEEE Trans. Veh. Technol.}, 73(9): 12556--12569.

\bibitem[{Liu et~al.(2024{\natexlab{b}})Liu, Tian, Zhao, Yu, Xie, Wang, Ye, Jiao, and Liu}]{liuVMambaVisualState2024a}
Liu, Y.; Tian, Y.; Zhao, Y.; Yu, H.; Xie, L.; Wang, Y.; Ye, Q.; Jiao, J.; and Liu, Y. 2024{\natexlab{b}}.
\newblock {{VMamba}}: Visual State Space Model.
\newblock In \emph{Advances in {{Neural Information Processing Systems}} 38: {{Annual Conference}} on {{Neural Information Processing Systems}} 2024, {{Neurips}} 2024, {{Vancouver}}, {{BC}}, {{Canada}}, {{December}} 10 - 15, 2024}.

\bibitem[{Liu et~al.(2020{\natexlab{a}})Liu, Tian, Glaser, and Kira}]{liuWhen2comMultiAgentPerception2020}
Liu, Y.-C.; Tian, J.; Glaser, N.; and Kira, Z. 2020{\natexlab{a}}.
\newblock When2com: {{Multi-Agent Perception}} via {{Communication Graph Grouping}}.
\newblock In \emph{2020 {{IEEE}}/{{CVF Conference}} on {{Computer Vision}} and {{Pattern Recognition}} ({{CVPR}})}, 4105--4114.

\bibitem[{Liu et~al.(2020{\natexlab{b}})Liu, Tian, Ma, Glaser, Kuo, and Kira}]{liuWho2comCollaborativePerception2020}
Liu, Y.-C.; Tian, J.; Ma, C.-Y.; Glaser, N.; Kuo, C.-W.; and Kira, Z. 2020{\natexlab{b}}.
\newblock Who2com: Collaborative Perception via Learnable Handshake Communication.
\newblock \emph{2020 IEEE International Conference on Robotics and Automation (ICRA)}, 6876--6883.

\bibitem[{Mao et~al.(2023)Mao, Shi, Wang, and Li}]{mao3DObjectDetection2023}
Mao, J.; Shi, S.; Wang, X.; and Li, H. 2023.
\newblock {{3D}} Object Detection for Autonomous Driving: A Comprehensive Survey.
\newblock \emph{Int. J. Comput. Vis.}, 131(8): 1909--1963.

\bibitem[{Park et~al.(2024)Park, Kim, Ko, Kim, and Kim}]{parkVideoMambaSpatiotemporalSelective2024}
Park, J.; Kim, H.-S.; Ko, K.; Kim, M.; and Kim, C. 2024.
\newblock {{VideoMamba}}: Spatio-Temporal Selective State Space Model.
\newblock In \emph{European {{Conference}} on {{Computer Vision}}, {{ECCV}} 2024}, volume 15083 of \emph{Lecture {{Notes}} in {{Computer Science}}}, 1--18.

\bibitem[{Phung et~al.(2024)Phung, Dao, Dao, Phan, Metaxas, and Tran}]{phungDiMSUMDiffusionMamba2024}
Phung, H.; Dao, Q.; Dao, T.~T.; Phan, V.~H.; Metaxas, D.~N.; and Tran, A.~T. 2024.
\newblock {{DiMSUM}}: Diffusion Mamba - a Scalable and Unified Spatial-Frequency Method for Image Generation.
\newblock In \emph{Advances in {{Neural Information Processing Systems}} 38: {{Annual Conference}} on {{Neural Information Processing Systems}} 2024, {{Neurips}} 2024, {{Vancouver}}, {{BC}}, {{Canada}}, {{December}} 10 - 15, 2024}.

\bibitem[{Qi et~al.(2017)Qi, Su, Mo, and Guibas}]{qiPointNetDeepLearning2017}
Qi, C.~R.; Su, H.; Mo, K.; and Guibas, L.~J. 2017.
\newblock {{PointNet}}: Deep Learning on Point Sets for {{3D}} Classification and Segmentation.
\newblock In \emph{2017 {{IEEE Conference}} on {{Computer Vision}} and {{Pattern Recognition}}, {{CVPR}} 2017, {{Honolulu}}, {{HI}}, {{USA}}, {{July}} 21-26, 2017}, 77--85.

\bibitem[{Qu et~al.(2024)Qu, Chen, Bai, Lu, Fan, Zhang, Fu, and Yang}]{quSiCPSimultaneousIndividual2024}
Qu, D.; Chen, Q.; Bai, T.; Lu, H.; Fan, H.; Zhang, H.; Fu, S.; and Yang, Q. 2024.
\newblock {{SiCP}}: Simultaneous Individual and Cooperative Perception for {{3D}} Object Detection in Connected and Automated Vehicles.
\newblock In \emph{2024 {{IEEE}}/{{RSJ International Conference}} on {{Intelligent Robots}} and {{Systems}} ({{IROS}})}, 8905--8912.

\bibitem[{Shi et~al.(2025)Shi, Xia, Jin, Wang, Zhao, Xia, Xiao, and Yang}]{shiVmambaIRVisualState2025}
Shi, Y.; Xia, B.; Jin, X.; Wang, X.; Zhao, T.; Xia, X.; Xiao, X.; and Yang, W. 2025.
\newblock {{VmambaIR}}: Visual State Space Model for Image Restoration.
\newblock \emph{IEEE Trans. Circuits Syst. Video Technol.}, 35(6): 5560--5574.

\bibitem[{Sun et~al.(2024)Sun, Wang, Su, and Deng}]{sunFrequencyawareNaturalCamouflage2024}
Sun, Q.; Wang, X.; Su, R.; and Deng, Y. 2024.
\newblock Frequency-Aware Natural Camouflage Object Segmentation.
\newblock In \emph{Sixth {{Conference}} on {{Frontiers}} in {{Optical Imaging}} and {{Technology}}: {{Imaging Detection}} and {{Target Recognition}}}, volume 13156, 263--274.

\bibitem[{Tan et~al.(2024)Tan, Zhao, Wei, Gu, Liu, and Wei}]{tanFrequencyawareDeepfakeDetection2024a}
Tan, C.; Zhao, Y.; Wei, S.; Gu, G.; Liu, P.; and Wei, Y. 2024.
\newblock Frequency-Aware Deepfake Detection: Improving Generalizability through Frequency Space Domain Learning.
\newblock In \emph{Thirty-Eighth {{AAAI Conference}} on {{Artificial Intelligence}}, {{AAAI}} 2024, {{Thirty-sixth Conference}} on {{Innovative Applications}} of {{Artificial Intelligence}}, {{IAAI}} 2024, {{Fourteenth Symposium}} on {{Educational Advances}} in {{Artificial Intelligence}}, {{EAAI}} 2014, {{February}} 20-27, 2024, {{Vancouver}}, {{Canada}}}, 5052--5060.

\bibitem[{Wang et~al.(2020)Wang, Manivasagam, Liang, Yang, Zeng, and Urtasun}]{wangV2VNetVehicletoVehicleCommunication2020}
Wang, T.-H.; Manivasagam, S.; Liang, M.; Yang, B.; Zeng, W.; and Urtasun, R. 2020.
\newblock {{V2VNet}}: {{Vehicle-to-Vehicle Communication}} for {{Joint Perception}} and {{Prediction}}.
\newblock In \emph{European {{Conference}} on {{Computer Vision}}, {{ECCV}} 2020}, volume 12347, 605--621.

\bibitem[{Wang et~al.(2024)Wang, Wang, Ding, Li, Wu, Rong, Kong, Huang, Li, Yang, Wang, Jiang, Li, Wang, Tian, and Tang}]{wangStateSpaceModel2024}
Wang, X.; Wang, S.; Ding, Y.; Li, Y.; Wu, W.; Rong, Y.; Kong, W.; Huang, J.; Li, S.; Yang, H.; Wang, Z.; Jiang, B.; Li, C.; Wang, Y.; Tian, Y.; and Tang, J. 2024.
\newblock State Space Model for New-Generation Network Alternative to Transformers: A Survey.
\newblock \emph{CoRR}, abs/2404.9516.

\bibitem[{Wu et~al.(2024)Wu, Jiang, Wang, Liu, Liu, Qiao, Ouyang, He, and Zhao}]{wuPointTransformerV32024}
Wu, X.; Jiang, L.; Wang, P.-S.; Liu, Z.; Liu, X.; Qiao, Y.; Ouyang, W.; He, T.; and Zhao, H. 2024.
\newblock Point Transformer {{V3}}: Simpler, Faster, Stronger.
\newblock In \emph{{{IEEE}}/{{CVF Conference}} on {{Computer Vision}} and {{Pattern Recognition}}, {{CVPR}} 2024, {{Seattle}}, {{WA}}, {{USA}}, {{June}} 16-22, 2024}, 4840--4851.

\bibitem[{Xu et~al.(2025)Xu, Zhang, Cai, and Huang}]{xuCoSDHCommunicationefficientCollaborative2025}
Xu, J.; Zhang, Y.; Cai, Z.; and Huang, D. 2025.
\newblock {{CoSDH}}: Communication-Efficient Collaborative Perception via Supply-Demand Awareness and Intermediate-Late Hybridization.
\newblock In \emph{{{IEEE}}/{{CVF Conference}} on {{Computer Vision}} and {{Pattern Recognition}}, {{CVPR}} 2025, {{Nashville}}, {{TN}}, {{USA}}, {{June}} 11-15, 2025}, 6834--6843.

\bibitem[{Xu et~al.(2021)Xu, Guo, Han, Xia, Xiang, and Ma}]{xuOpenCDAOpenCooperative2021}
Xu, R.; Guo, Y.; Han, X.; Xia, X.; Xiang, H.; and Ma, J. 2021.
\newblock {{OpenCDA}}: {{An Open Cooperative Driving Automation Framework Integrated}} with {{Co-Simulation}}.
\newblock In \emph{2021 {{IEEE International Intelligent Transportation Systems Conference}} ({{ITSC}})}, 1155--1162.

\bibitem[{Xu et~al.(2022{\natexlab{a}})Xu, Tu, Xiang, Shao, Zhou, and Ma}]{xuCoBEVTCooperativeBirds2022}
Xu, R.; Tu, Z.; Xiang, H.; Shao, W.; Zhou, B.; and Ma, J. 2022{\natexlab{a}}.
\newblock {{CoBEVT}}: Cooperative Bird's Eye View Semantic Segmentation with Sparse Transformers.
\newblock In \emph{Conference on {{Robot Learning}}, {{Corl}} 2022, 14-18 {{December}} 2022, {{Auckland}}, {{New Zealand}}}, volume 205 of \emph{Proceedings of {{Machine Learning Research}}}, 989--1000.

\bibitem[{Xu et~al.(2023)Xu, Xia, Li, Li, Zhang, Tu, Meng, Xiang, Dong, Song, Yu, Zhou, and Ma}]{xuV2V4RealRealWorldLargeScale2023a}
Xu, R.; Xia, X.; Li, J.; Li, H.; Zhang, S.; Tu, Z.; Meng, Z.; Xiang, H.; Dong, X.; Song, R.; Yu, H.; Zhou, B.; and Ma, J. 2023.
\newblock {{V2V4Real}}: {{A Real-World Large-Scale Dataset}} for {{Vehicle-to-Vehicle Cooperative Perception}}.
\newblock In \emph{2023 {{IEEE}}/{{CVF Conference}} on {{Computer Vision}} and {{Pattern Recognition}} ({{CVPR}})}, 13712--13722.

\bibitem[{Xu et~al.(2022{\natexlab{b}})Xu, Xiang, Tu, Xia, Yang, and Ma}]{xuV2XViTVehicletoeverythingCooperative2022}
Xu, R.; Xiang, H.; Tu, Z.; Xia, X.; Yang, M.-H.; and Ma, J. 2022{\natexlab{b}}.
\newblock {{V2X-ViT}}: Vehicle-to-Everything Cooperative Perception with~Vision Transformer.
\newblock In \emph{European {{Conference}} on {{Computer Vision}}, {{ECCV}} 2022}, volume 13699 of \emph{Lecture {{Notes}} in {{Computer Science}}}, 107--124.

\bibitem[{Xu et~al.(2022{\natexlab{c}})Xu, Xiang, Xia, Han, Li, and Ma}]{xuOPV2VOpenBenchmark2022a}
Xu, R.; Xiang, H.; Xia, X.; Han, X.; Li, J.; and Ma, J. 2022{\natexlab{c}}.
\newblock {{OPV2V}}: {{An Open Benchmark Dataset}} and {{Fusion Pipeline}} for {{Perception}} with {{Vehicle-to-Vehicle Communication}}.
\newblock In \emph{2022 {{International Conference}} on {{Robotics}} and {{Automation}} ({{ICRA}})}, 2583--2589.

\bibitem[{Xu et~al.(2024)Xu, Yang, Wang, Cai, Du, and Chen}]{xuVisualMambaSurvey2024}
Xu, R.; Yang, S.; Wang, Y.; Cai, Y.; Du, B.; and Chen, H. 2024.
\newblock Visual Mamba: A Survey and New Outlooks.
\newblock arXiv:2404.18861.

\bibitem[{Yang et~al.(2023)Yang, Yang, Wang, Liu, Xu, Yin, Zhai, and Zhang}]{yangHow2commCommunicationEfficientCollaborationPragmatic2023}
Yang, D.; Yang, K.; Wang, Y.; Liu, J.; Xu, Z.; Yin, R.; Zhai, P.; and Zhang, L. 2023.
\newblock How2comm: {{Communication-Efficient}} and {{Collaboration-Pragmatic Multi-Agent Perception}}.
\newblock \emph{Advances in Neural Information Processing Systems}, 36: 25151--25164.

\bibitem[{Yao et~al.(2025)Yao, Liu, Cui, Peng, and Zhou}]{yaoSelectiveVisualPrompting2025}
Yao, Y.; Liu, Z.; Cui, Z.; Peng, Y.; and Zhou, J. 2025.
\newblock Selective Visual Prompting in Vision Mamba.
\newblock In \emph{{{AAAI-25}}, {{Sponsored}} by the {{Association}} for the {{Advancement}} of {{Artificial Intelligence}}, {{February}} 25 - {{March}} 4, 2025, {{Philadelphia}}, {{PA}}, {{USA}}}, 22083--22091.

\bibitem[{Yazgan et~al.(2024)Yazgan, Graf, Liu, Fleck, and Z{\"o}llner}]{yazganSurveyIntermediateFusion2024}
Yazgan, M.; Graf, T.; Liu, M.; Fleck, T.; and Z{\"o}llner, J.~M. 2024.
\newblock A Survey on Intermediate Fusion Methods for Collaborative Perception Categorized by Real World Challenges.
\newblock In \emph{{{IEEE Intelligent Vehicles}}, {{June}} 2-5, 2024}, 2226--2233.

\bibitem[{Yu et~al.(2022)Yu, Luo, Shu, Huo, Yang, Shi, Guo, Li, Hu, Yuan, and Nie}]{yuDAIRV2XLargeScaleDataset2022}
Yu, H.; Luo, Y.; Shu, M.; Huo, Y.; Yang, Z.; Shi, Y.; Guo, Z.; Li, H.; Hu, X.; Yuan, J.; and Nie, Z. 2022.
\newblock {{DAIR-V2X}}: {{A Large-Scale Dataset}} for {{Vehicle-Infrastructure Cooperative 3D Object Detection}}.
\newblock In \emph{2022 {{IEEE}}/{{CVF Conference}} on {{Computer Vision}} and {{Pattern Recognition}} ({{CVPR}})}, 21329--21338.

\bibitem[{Zhang et~al.(2024)Zhang, Fan, He, Lei, Zhang, and Zhang}]{zhangVoxelMambaGroupfree2024}
Zhang, G.; Fan, L.; He, C.; Lei, Z.; Zhang, Z.; and Zhang, L. 2024.
\newblock Voxel Mamba: Group-Free State Space Models for Point Cloud Based {{3D}} Object Detection.
\newblock In \emph{Advances in {{Neural Information Processing Systems}} 38: {{Annual Conference}} on {{Neural Information Processing Systems}} 2024, {{Neurips}} 2024, {{Vancouver}}, {{BC}}, {{Canada}}, {{December}} 10 - 15, 2024}.

\bibitem[{Zhong et~al.(2022)Zhong, Li, Tang, Kuang, Wu, and Ding}]{zhongDetectingCamouflagedObject2022}
Zhong, Y.; Li, B.; Tang, L.; Kuang, S.; Wu, S.; and Ding, S. 2022.
\newblock Detecting Camouflaged Object in Frequency Domain.
\newblock In \emph{2022 {{IEEE}}/{{CVF Conference}} on {{Computer Vision}} and {{Pattern Recognition}} ({{CVPR}})}, 4494--4503.

\bibitem[{Zhou and Tuzel(2018)}]{zhouVoxelNetEndtoendLearning2018}
Zhou, Y.; and Tuzel, O. 2018.
\newblock {{VoxelNet}}: End-to-End Learning for Point Cloud Based {{3D}} Object Detection.
\newblock In \emph{2018 {{IEEE}}/{{CVF Conference}} on {{Computer Vision}} and {{Pattern Recognition}}}, 4490--4499.

\bibitem[{Zhu et~al.(2024)Zhu, Liao, Zhang, Wang, Liu, and Wang}]{zhuVisionMambaEfficient2024}
Zhu, L.; Liao, B.; Zhang, Q.; Wang, X.; Liu, W.; and Wang, X. 2024.
\newblock Vision Mamba: Efficient Visual Representation Learning with Bidirectional State Space Model.
\newblock In \emph{Forty-First {{International Conference}} on {{Machine Learning}}, {{ICML}} 2024, {{Vienna}}, {{Austria}}, {{July}} 21-27, 2024}.

\bibitem[{Zimerman, Ali, and Wolf(2024)}]{zimermanExplainingModernGatedlinear2024}
Zimerman, I.; Ali, A.; and Wolf, L. 2024.
\newblock Explaining Modern Gated-Linear {{RNNs}} via a Unified Implicit Attention Formulation.
\newblock arXiv:2405.16504.

\bibitem[{Zou et~al.(2024)Zou, Yu, Huang, and Zhao}]{zouFreqMambaViewingMamba2024}
Zou, Z.; Yu, H.; Huang, J.; and Zhao, F. 2024.
\newblock {{FreqMamba}}: Viewing Mamba from a Frequency Perspective for Image Deraining.
\newblock In \emph{Proceedings of the 32nd {{ACM International Conference}} on {{Multimedia}}}, {{MM}} '24, 1905--1914.

\end{thebibliography}


%
%

\ifshowappendix
\newpage
\appendix

\twocolumn[%
  \begin{center}
    {\LARGE\bfseries Supplementary Material\par}
  \end{center}
  \vspace{3em}
]

\section{Design Rationale and Key Highlights of our \ourname}

This work aims to develop a compact 3D scene representation that encodes both rich structural detail and explicit semantic cues as an intermediate feature for collaborative perception, thereby enabling more efficient computation and communication. 

Traditional 2D BEV feature maps discard vertical height information—crucial for distinguishing object categories (e.g., cars, pedestrians, bicycles, roadways, and trees exhibit substantially different height characteristics), revealing terrain variations, and delineating object boundaries—and suffer further accuracy loss when subjected to additional channel compression or spatial filtering. 

Moreover, these methods maintain dense 2D grids throughout feature extraction and fusion, forcing redundant processing of vast background and empty regions; Multi-agent fusion must enlarge these grids along both spatial axes to encompass all agents' viewpoints, resulting in significant redundant computational overhead.
Ideally, computation should focus solely on object-related features, with the ability to dynamically adjust workload of  computational cost—similar to Where2comm’s adaptive communication—so that, under tight resource budgets, only the most informative tokens are processed.

Simultaneously, the emergence of large-scale sequence models in vision and language underscores the necessity for a scalable, high-capacity collaborative perception framework. The majority of backbone networks in current collaborative frameworks are either convolutional neural networks (CNNs) or transformers. However, CNNs face challenges in scaling model size and parameter count to develop more robust general models. Furthermore, certain transformer-based approaches operate on 2D feature maps with compressed structural information, often by forming patches or dividing windows, which results in cumulative information loss. There is a desire to directly represent 3D environments as 1D sequences, thereby retaining as much detailed information as possible for input into large-scale, general sequence processing models. 

Moreover, future frameworks should generate comprehensive, compact, and reliable intermediate features that seamlessly and efficiently interface with large language models (LLMs) or vision–language models (VLMs), aligning with emerging technological trends. When adapting representations for LLM integration, directly converting raw 3D features into sequential inputs is more elegant than first compressing 3D scenes into 2D feature maps and subsequently into 1D patch sequences, as the latter multi-stage transformation inevitably accumulates information loss.

This paper proposes leveraging 3D point-level tokens as intermediate features for collaborative perception, overcoming inherent limitations and shortcomings of traditional 2D BEV-based frameworks. Our method produces a one-dimensional token sequence that preserves full spatial structure and clear semantics in a highly compact form, supports dynamic sequence-length adjustment to minimize both computation and communication costs, and is fully compatible with emerging large-scale sequence-processing architectures.

We believe that leveraging 3D point-level tokens for collaborative perception represents a promising direction for future research. We hope this work will advance the field and accelerate the real-world deployment of efficient, reliable autonomous systems.

\section{Limitations and Future Works}

The \ourname  proposed in this study addresses the inherent limitations of traditional 2D BEV based collaborative perception. It facilitates efficient and dynamic computation and communication for multi-agent collaboration. Extensive experiments conducted across multiple datasets validate that \ourname achieves an effective balance between perception performance and computational-communication efficiency. Nevertheless, several limitations require further exploration, as outlined below:

\textbf{Independent Token Selection.} Currently, \ourname selects the top-$k$ tokens based on either communication or computation budget constraints independently, without \textit{a joint optimization model that integrates communication costs, computation costs, and perception performance}. Ideally, a neighboring agent could dynamically determine an optimal length for the point-level token sequence by considering its own computational resources, those of the ego agent, and the quality of communication between them. Furthermore, each agent can dynamically adjust its token processing load to optimize its computational resource usage.

\textbf{Static Top-$k$ Threshold.} The threshold for selecting the top-$k$ tokens remains static and does not adapt to scene complexity or task-specific demands. In highly complex collaborative scenarios, this fixed ratio may overlook critical details necessary for downstream tasks; whereas in simpler scenes, it may retain redundant tokens. A dynamic, scene- and task-specific top-$k$ selection strategy is therefore essential.

\textbf{Suboptimal Detection Head.} For experimental fairness, this study employs a 2D convolutional detection head for 3D detection tasks, aligning with existing methods. While this framework preserves comprehensive 3D structural information until the pipeline’s final stages, its integration with the task network remains inelegant. Directly generating 3D detection proposals from point-level tokens could provide a more seamless solution.

\textbf{Untapped Potential for Multiple Complex Tasks.} Given that point-level tokens encapsulate detailed structural and rich semantic information, they possess latent capabilities for addressing multiple complex tasks. Their integration with more sophisticated 3D perception tasks represents a promising area for future research.

\textbf{Limited Modality Support.} The current framework relies solely on LiDAR point cloud inputs for collaborative perception. Incorporating features from additional modalities, such as camera data, poses a practical challenge that warrants thorough investigation.

\textbf{Adaptation to Large Language Models.} Adapting point-level token sequences for compatibility with LLMs emerges as a valuable and forward-looking research direction.

This study marks an initial step toward unlocking the potential of robust, trustworthy, and reliable point-level tokens for collaborative perception. The development and deployment of the \ourname framework in real-world applications constitute a promising and meaningful trajectory for future work.


\section{Experimental Settings}
\label{experiment}

\subsection{Datasets}
\label{data}

To thoroughly assess the effectiveness of our method in both simulated and real-world environments, we conducted extensive experiments on three representative datasets encompassing both simulation and real-world scenarios:

OPV2V \cite{xuOPV2VOpenBenchmark2022a} is a large-scale simulated dataset for multi-agent V2V perception, co-simulated with Carla \cite{dosovitskiyCARLAOpenUrban2017} and OpenCDA \cite{xuOpenCDAOpenCooperative2021}. The dataset features 73 scenes across 6 road types in 9 cities, comprising 12K LiDAR point cloud frames and 230K annotated 3D bounding boxes. It includes a total of 10,914 3D annotated LiDAR frames, divided into training (6,764 frames), validation (1,981 frames), and testing (2,169 frames) sets.

V2V4Real \cite{xuV2V4RealRealWorldLargeScale2023a} is a large-scale real-world dataset for vehicle-to-vehicle (V2V) collaborative perception, aimed at advancing autonomous driving technologies. It features data from two vehicles equipped with multimodal sensors across various scenarios, including 20,000 LiDAR point cloud frames, 40,000 RGB images, and 240,000 annotated 3D bounding boxes, covering a perception range of 280m×80m over 410 km.

DAIR-V2X \cite{yuDAIRV2XLargeScaleDataset2022} is a real-world dataset for collaborative perception in V2X scenarios, consisting of 71,254 frames collected from a vehicle and a roadside unit (RSU), both equipped with LiDAR and a 1920×1080 camera. The vehicle's LiDAR has 40 channels, while the RSU's LiDAR features 300 channels, providing complementary perspectives for cooperative perception tasks.

\subsection{Implementations}

The proposed models are implemented in PyTorch 2.3 and trained on an NVIDIA RTX 4090 using the Adam optimizer \cite{kingmaAdamMethodStochastic2015}. The initial learning rate is set to $1\times10^{-3}$ and decayed by a factor of 0.1 every ten epochs, for a total of 40 epochs. We utilize smooth $L_1$ loss for regression and focal loss for classification, consistent with existing baseline methods \cite{huWhere2commCommunicationEfficientCollaborative2022b, liCollaMambaEfficientCollaborative2024, xuV2XViTVehicletoeverythingCooperative2022}. Experimental setups follow the perception range specifications in \mytab{table:range-settings}. During training, each model is supervised on the training split, with the epoch yielding the lowest validation loss used to save model weights; final evaluation is conducted on the test split. These settings are applied uniformly across all baseline methods.

\begin{table}[ht]
  \centering
  \caption{Perception ranges for each dataset.}
  \label{table:range-settings}
  \small
  \begin{tabular}{@{}lccc@{}}
    \toprule
    Range      & x               & y               & z       \\
    \midrule
    OPV2V      & $[-140,140]$    & $[-40,40]$      & $[-3,1]$ \\
    V2V4Real   & $[-102.4,102.4]$& $[-51.2,51.2]$  & $[-3,1]$ \\
    DAIR-V2X   & $[-100.8,100.8]$& $[-40,40]$      & $[-3,1]$ \\
    \bottomrule
  \end{tabular}
\end{table}

\begin{table*}[ht]
  \centering
  \caption{Inference time, computational load, and memory usage for different models on OPV2V.}
  \label{table:profiling}
  \small
  \begin{tabular}{@{}l S[table-format=1.3] S[table-format=1.3] S[table-format=3.3] S[table-format=1.3] S[table-format=2.3]@{}}
    \toprule
    Model              & {CPU Time (s)} & {CUDA Time (s)} & {GFLOPs}     & {Peak GPU Mem (GB)} & {Params (M)} \\
    \midrule
    AttFuse            & 0.02762        & 0.04334         & 188.101      & 0.406               & 8.058       \\
    V2VNet             & 0.11395        & 0.16282         & 880.813      & 0.529               & 14.613      \\
    V2X-ViT            & 0.25912        & 0.28181         & 545.085      & 1.276               & 13.453      \\
    CoBEVT             & 0.08884        & 0.20085         & 402.078      & 1.301               & 10.514      \\
    Where2comm         & 0.05352        & 0.08458         & 420.074      & 0.517               & 11.429      \\
    SiCP               & 0.04848        & 0.06543         & 277.716      & 0.863               & 7.403       \\
    CoSDH              & 0.08182        & 0.12810         & 796.648      & 0.549               & 14.453      \\
    CollaMamba         & 0.15324        & 0.16366         & 198.260      & 0.921               & 3.920       \\
    CoMamba            & 0.03742        & 0.07437         & 238.794      & 0.991               & 9.219       \\
    \rowcolor[HTML]{EFEFEF}
    \ourname              & 0.33619        & 0.25234         & 176.210      & 0.578               & 6.368       \\
    \bottomrule
  \end{tabular}
\end{table*}

\subsection{Training Details}
When training our \ourname model, an end-to-end training strategy could involve simultaneously optimizing all components, including the encoder, fusion blocks, token semantic importance filtering block, and neighbor-to-ego alignment block. However, this approach presents a challenge: early in training, insufficiently trained encoder blocks may generate token sequences with inaccurate semantic features, which can mislead the semantic importance filtering process. Consequently, the selected tokens may not reliably represent object-related foreground elements, leading to errors that propagate through the fusion and neighbor-to-ego alignment blocks, accumulating towards the end of the pipeline and slowing overall model convergence. To address this issue, we propose a three-stage pretraining-finetuning strategy: (1) Set a large top-$k$ threshold for semantic importance filtering to retain more tokens, enabling the encoder backbone to develop robust semantic modeling capabilities. (2) Reduce the top-$k$ threshold to select object-related foreground tokens for finetuning, leveraging the backbone's enhanced semantic modeling to accurately distinguish foreground from background while optimizing the fusion blocks for effective foreground relationship modeling. (3) Train the neighbor-to-ego alignment block independently in noisy environments to enhance its alignment and correction capabilities.

\section{More Details of \ourname}

\subsection{Model Efficiency}

In \mytab{table:profiling}, we report efficiency metrics measured with Torch Profiler
\footnote{https://docs.pytorch.org/docs/stable/profiler.html} 
on the OPV2V test dataset using a single NVIDIA RTX 4090 and a batch size of 1. Metrics include CPU time (s), CUDA time (s), computational cost (GFLOPs), peak GPU memory usage (GB), and model size (M parameters). \ourname leads across all metrics. Its CPU time is about 0.07 s higher than that of V2X-ViT, attributable to additional sorting and indexing operations, but this overhead remains within acceptable limits and suggests a direction for future optimization. Note that GPU memory usage reported by Torch Profiler tends to understate the consumption observed via ``nvidia-smi" command; real-world deployment should include comprehensive profiling in the target environment.

\subsection{About Point Tokenizer}


The point tokenizer proposed in this study is designed to transform raw point clouds into point-level token feature representations. Raw point clouds frequently exhibit uneven spatial distribution, substantial redundancy due to their large volume, and susceptibility to noise. Traditional sampling techniques, such as Farthest Point Sampling (FPS) \cite{qiPointNetDeepLearning2017,hanMamba3DEnhancingLocal2024,liangPointMambaSimpleState2024}, retain the original uneven distribution and impose significant computational burdens, making them impractical for outdoor point cloud applications. Similarly, sparse voxel sampling, widely adopted in frameworks like OpenCOOD
\footnote{https://github.com/DerrickXuNu/OpenCOOD}, 
retains only a limited subset of points within each voxel grid and aggregates their features by selecting the maximum feature value, which sacrifices considerable spatial and structural information \cite{xuCoSDHCommunicationefficientCollaborative2025, huWhere2commCommunicationEfficientCollaborative2022b, quSiCPSimultaneousIndividual2024}.

In contrast, our point tokenizer employs a grid-based approach, initiating the process by positioning sampling points at predefined intervals within the 3D space. A sampling point is converted into a token if a sufficient number of raw points reside within its designated grid interval. Feature aggregation entails calculating a range of statistics for these points, including mean coordinates, dispersion (the average absolute distance from each point to the mean), density (the number of points normalized by a fixed value), grid offset (the average absolute distance to the sampling point), and intensity metrics (mean, maximum, and standard deviation), alongside the distance to the sensor. Operationally, the normalized coordinates of raw points are divided by the grid interval and floored, ensuring that points within the same grid cell share identical grid coordinates. These points’ features are then aggregated into a single point-level token feature, and this process incurs minimal computational complexity. The resulting tokens encapsulate rich spatial characteristics, facilitating noise filtering and object detection. For example, sparse, disordered point distributions around a grid point likely signify noise, whereas dense, structured distributions with distinct intensity profiles at greater distances from the sensor suggest object-related foreground elements.

Comparative experiments, detailed in \mytab{table:sparse-tokenizer-comparison}, substantiate the efficacy of our point tokenizer, revealing improved perception performance over sparse voxel methods, with notably greater enhancements on real-world datasets. This superiority arises from the tokenizer’s capacity to comprehensively capture the complex spatial distributions and structural nuances prevalent especially in real-world point clouds.

\mytab{table:grid-interval} illustrates how varying the grid interval during sampling affects perception performance. As shown, a larger grid interval degrades perception accuracy, since aggregating features from overly large spatial regions leads to a loss of detailed structural information. Conversely, a smaller grid interval results in excessively high training costs, making it impractical for real-world deployment. After balancing perception performance against computational overhead, we select a grid interval of 0.4 for all subsequent comparative experiments in this study. Additionally, this choice aligns with the voxel-based sampling intervals used by baseline methods, ensuring fairness in performance comparisons.

\begin{table}[!ht]
  \centering
  \caption{Comparison of detection performance among sparse‐voxel baseline and our point tokenizer.}
  \label{table:sparse-tokenizer-comparison}
  \small
  \begin{tabular}{@{}l c c c@{}}
    \toprule
    \multirow{2}{*}{Method}
      & \multicolumn{1}{c}{OPV2V}
      & \multicolumn{1}{c}{V2V4Real}
      & \multicolumn{1}{c}{DAIR-V2X} \\ 
    & {AP@0.5/0.7} & {AP@0.5/0.7} & {AP@0.5/0.7} \\
    \midrule
    Sparse Voxel            & 0.971 / 0.928 & 0.639 / 0.413 & 0.760 / 0.541 \\
    \rowcolor[HTML]{EFEFEF}
    Point Tokenizer         & 0.973 / 0.934 & 0.644 / 0.447 & 0.761 / 0.593 \\
    \bottomrule
  \end{tabular}
\end{table}

\begin{table}[ht]
  \centering
  \caption{Effect of grid interval on perception performance and computational overhead.}
  \label{table:grid-interval}
  \small
  \begin{tabular}{@{}l c c c@{}}
    \toprule
    Grid interval       & \shortstack{OPV2V\\AP@0.5/0.7} & GFLOPs   & \shortstack{Peak GPU\\Memory (GB)} \\ 
    \midrule
    0.1                 & 0.965 / 0.912                 & 1222.540 & 6.095  \\
    0.2                 & 0.973 / 0.933                 & 716.713  & 3.124  \\
    \rowcolor[HTML]{EFEFEF}
    0.4—\ourname            & 0.973 / 0.934                 & 176.210  & 0.578  \\
    0.8                 & 0.946 / 0.870                 & 51.218   & 0.546  \\
    1.6                 & 0.900 / 0.805                 & 14.612   & 0.257  \\
    \bottomrule
  \end{tabular}
\end{table}

\subsection{About frequency-enhanced state space model}
\label{suppl:fssm}

The Mamba framework excels at efficient long-range modeling on two-dimensional images and on compact  point-cloud sequences\cite{zhuVisionMambaEfficient2024,guEfficientMixedHeterogeneous2025,jinUniMambaUnifiedSpatialchannel2025}, but its direct deployment in wide-area outdoor LiDAR scenes remains sub-optimal. In these environments, foreground vehicles occupy only a minute fraction of the point cloud and their spatial signatures often resemble the amorphous patterns produced by roads, vegetation,..., and sensor noise. Vehicle points blend into the background, exhibiting fuzzy boundaries that undermine the discriminatory power of embeddings based solely on raw coordinates. Consequently, the gating operations in the selective state-space model tend to suppress informative vehicle evidence along with clutter.

Biological vision suggests an alternative cue. Psychophysical and ecological studies indicate that humans and many predators rely on frequency-selective filters to extract camouflaged targets from complex backgrounds, likely implying an implicit, rapid Fourier-like analysis that accentuates shape-specific spectral components \cite{baumbachPsychophysicsHumanVision2010, sunFrequencyawareNaturalCamouflage2024, dasPredictingHumanCamouflage2023, dasCamouflageDetectionExperiments2022, acklehFrequencydependentEvolutionPredator2021}. For point-cloud perception, a compact spectral descriptor can highlight the narrow-band contours characteristic of vehicles while attenuating broadband terrain signatures. Moreover, after fusing point-level tokens from multiple vehicles, frequency-domain representations can unveil deeper intrinsic correlations and further amplify the salience of object-related features.

The mathematical correspondence between the selective state space model and linear attention provides a natural insertion point for this descriptor \cite{hanDemystifyMambaVision2024, aliHiddenAttentionMamba2024, zimermanExplainingModernGatedlinear2024}. The output matrix $C$ in the state-space formulation serves the same “query” role as the matrix $Q$ in attention, so augmenting $C$ with a frequency prompt injects biologically inspired prior knowledge without altering Mamba’s linear-time recurrent dynamics. This frequency-aware query helps the network preserve subtle vehicle features and improves its ability to separate foreground tokens from extensive background clutter in outdoor LiDAR data.

Building on these insights, we endow Mamba with frequency-aware auxiliary cues that emulate biological vision and propose the frequency-enhanced state space model, enabling a more precise analysis of object-related semantics. We next examine the underlying similarities between Mamba and attention mechanisms.

\textbf{State space model}. The classical state space model (SSM) describes a continuous‐time latent trajectory $h(t)\in\mathbb{R}^{d}$ that evolves under linear dynamics and is driven by an external signal $x(t)\in\mathbb{R}$:


\begin{equation}
\begin{aligned}
\dot{h}(t) &=Ah(t)+Bx(t), \\
y(t) &=Ch(t)+Dx(t),
\end{aligned}
\end{equation}
where $A\in\mathbb{R}^{d\times d}$ governs the intrinsic dynamics, $B\in\mathbb{R}^{d}$ injects the input, $C\in\mathbb{R}^{1\times d}$ projects the state to the output, and $D\in\mathbb{R}$ provides a direct input shortcut.  Using a zero-order hold with step size $\Delta$, the system is discretised as


\begin{subequations}
\begin{align}
\overline{A} &=\exp(\Delta A),  \\
\overline{B}
  &= (\Delta A)^{-1}\!\left(\exp(\Delta A)-I\right)\,\Delta B
  \;\approx\; \Delta B ,  \label{eq:approxB}
\end{align}
\end{subequations}
yielding the recurrent form:
\begin{equation}
\begin{aligned}
h_{i}  &= \overline{A}h_{i-1}+\overline{B}x_{i}, \\
y_{i}  &= Ch_{i}+Dx_{i}.
\end{aligned}
\end{equation}

Mamba augments the discrete SSM with input-dependent parameter selection.  For each token $x_{i}$ the network generates a token-specific timescale $\Delta_{i}$ together with an input matrix $B_{i}$ and an output matrix $C_{i}$, so that the effective system matrices $\overline{A}_{i},\overline{B}_{i},C_{i}$ are instantiated on the fly.  Thus, we obtain the following formulation:
\begin{equation}
\begin{aligned}
h_{i}&=\overline{A}_{i}h_{i-1}+\overline{B}_{i}x_{i},\\
y_{i}&=C_{i}h_{i}+Dx_{i}.
\end{aligned}
\end{equation}

In the above expression, by referring to \myeq{eq:approxB} and noting that $\overline{B}_{i}\,x_i \approx \Delta_{i} \overline{B}_{i} x_i = \overline{B}_{i}\bigl(\Delta_{i} x_i\bigr)$, we can rewrite the equation in the following general form:
\begin{equation}
\begin{aligned}
h_{i} &=\overline{A}_{i}h_{i-1}+\overline{B}_{i}\bigl(\Delta_{i} x_i\bigr),\\
y_{i} &=C_{i}h_{i} / 1 + Dx_{i},
\end{aligned}
\label{eq:ssm}
\end{equation}
here $\Delta_{i}$ functions as an input gate, $A_{i}$ acts as a forget gate, and the fixed residual term $Dx_{i}$ preserves a direct information path.  This selective SSM maintains the $O(N)$ runtime of the original model while allowing per-token adaptation of memory and read-out, which is crucial for handling long-range sequences such as point-cloud tokens in outdoor LiDAR scenes.

\textbf{Linear attention}.
Linear attention  mitigates the quadratic cost of standard self-attention by replacing the non-linear soft-max with a linear normalization and introducing a kernel function $\phi(\cdot)$. Given $Q=\phi(XW_Q),\;K=\phi(XW_K),\;V=XW_V$, we can derive:

\begin{equation}
\begin{aligned}
y_i
  = \sum_{j=1}^{N}
      \frac{Q_i K_j^{\top}}
           {\sum_{j=1}^{N} Q_i K_j^{\top}}\,
      V_j
  \;=\;
  \frac{Q_i\!\left(\sum_{j=1}^{N} K_j^{\top} V_j\right)}
       {Q_i\!\left(\sum_{j=1}^{N} K_j^{\top}\right)} .
\end{aligned}
\end{equation}

Here, the associative property of matrix multiplication permits the re-ordering
$
(QK^{\!\top})V = Q(K^{\!\top}V),
$
reducing the complexity from $O(N^{2})$ to $O(N)$.  In the full (non-causal) form each query attends to all keys and values.  For autoregressive settings the receptive field is restricted to past tokens, yielding the causal formulation:

\begin{equation}
\begin{aligned}
y_i=\frac{Q_i\sum_{j\le i}K_j^{\!\top}V_j}{Q_i\sum_{j\le i}K_j^{\!\top}}
       =\frac{Q_iS_i}{Q_iZ_i},
\end{aligned}
\end{equation}
where
\begin{equation}
\begin{aligned}
S_i &= \sum_{j\le i} K_j^{\top} V_j  = S_{i-1}+K_i^{\!\top}V_i,  \\
Z_i &= \sum_{j\le i} K_j^{\top} = Z_{i-1}+K_i^{\!\top}.
\end{aligned}
\end{equation}

This yields the following general recurrent linear-attention formulation:

\begin{equation}
\begin{aligned}
S_i &= \mathbf{1}\odot S_{i-1} + K_i^{\top}\bigl(\mathbf{1}\odot V_i\bigr),\\
y_i &= {Q_i S_i} / {Q_i Z_i} + \mathbf{0}\odot x_i. 
\end{aligned}
\label{eq:att}
\end{equation}

\textbf{Expected benefit.}
A close correspondence emerges upon comparing \myeq{eq:ssm} with \myeq{eq:att}. In particular, the following equivalences can be observed:
$h_i \leftrightarrow S_i \in \mathbb{R}^{d\times C}$,
$\overline{B}_i \leftrightarrow K_i^{\!\top} \in \mathbb{R}^{d\times 1}$,
$x_i \leftrightarrow V_i \in \mathbb{R}^{1\times C}$, and
$C_i \leftrightarrow Q_i \in \mathbb{R}^{1\times d}$.
Consequently, we further scrutinize the output matrix $C_i$ in \myeq{eq:ssm}, which functions analogously to the query vector in traditional attention mechanisms. By embedding frequency-domain features into $C_i$, $C_i$ can act as an auxiliary query that probes object-related characteristics of $x_i$ in the complex frequency domain, thereby enhancing the semantic expressiveness of $x_i$.


By granting the SSM a frequency-conditioned-liked query, $C \leftarrow C + \gamma Q_{freq}$, the network can not only retain the long-range, linear-time memory advantages of Mamba, but also exploit biologically motivated frequency filtering to isolate small, low-contrast vehicles from large swaths of background terrain points. 

\begin{equation}
\begin{aligned}
  h_i &= \overline{A}_i\,h_{i-1} + \overline{B}_i\bigl(\Delta x_i\bigr), \\
  y_i &= \Bigl(C_i + \gamma\,Q^{freq}_{i}\Bigr)\,h_i + D\,x_i,
\end{aligned}
\label{eq:sufssme}
\end{equation}
where $\gamma>0$ is a learnable scaling factor that balances the numerical range. 

Preliminary experiments on outdoor LiDAR benchmarks confirm that the augmented model achieves higher performance at negligible computational cost, demonstrating the practical relevance of frequency-enhanced state space reasoning.

\begin{subequations}\label{eq:freq}
\begin{align}
& X_{sc} = \operatorname{Norm}\bigl(\operatorname{Conv2D}(F_{sc})\bigr),
  \label{eq:Xsc} \\[2pt]
& (u_i, v_i)\;=\;\Bigl(\big\lfloor\tfrac{y_i}{s_{\mathrm{dx}}}\big\rfloor,\;
  \big\lfloor\tfrac{x_i}{s_{\mathrm{dy}}}\big\rfloor\Bigr),
  \quad i=1,\dots,M
  \label{eq:uv} \\[2pt]
& P_i(c,p,q)\;=\;X_{sc}\!\bigl(c,\;u_i + p - \tfrac{k-1}{2},\;
  v_i + q - \tfrac{k-1}{2}\bigr),
  \label{eq:patch} \\
& \quad p,q=0,\dots,k-1,\;\;c=1,\dots,C, \\[2pt]
& \widehat{P}_i(c,r,s) \;=\;\frac{1}{\sqrt{H_k W_k}}
  \sum_{p=0}^{H_k-1}\sum_{q=0}^{W_k-1}
  P_i(c,p,q)\,
  e^{-2\pi j\bigl(\tfrac{pr}{H_k}+\tfrac{qs}{W_k}\bigr)},
  \label{eq:dft} \\
& \quad r,s=0,\dots,k-1, \\[2pt]
& E_i^{\mathrm{DC}}(c)=\bigl\lvert\widehat{P}_i(c,0,0)\bigr\rvert,
  \label{eq:dc} \\
& E_i^{\mathrm{low}}(c)
  =\sum_{\substack{0<\sqrt{r^2+s^2}\le \alpha k}}
  \bigl\lvert\widehat{P}_i(c,r,s)\bigr\rvert,
  \label{eq:low} \\
& E_i^{\mathrm{high}}(c)
  =\sum_{\sqrt{r^2+s^2}>\beta k}
  \bigl\lvert\widehat{P}_i(c,r,s)\bigr\rvert,
  \label{eq:hi} \\
& S_i(c) =\frac{E_i^{\mathrm{high}}(c)}{E_i^{\mathrm{low}}(c)+\varepsilon},
  \label{eq:S} \\
& \nu_i = \bigl[E_i^{\mathrm{DC}},\;E_i^{\mathrm{low}},\;
  E_i^{\mathrm{high}},\;S_i\bigr],
  \label{eq:phi} \\
& \nu^{freq} = [\nu_i,\dots], \quad i=1,\dots,M,
  \label{eq:phifreq} \\
& Q^{freq} = \operatorname{Proj}\bigl(\nu^{freq}\bigr).
  \label{eq:proj}
\end{align}
\end{subequations}

\textbf{Frequency-domain features.}
To lighten the frequency-domain branch, we first downsample the dense scene feature map along the two spatial axes using strides $s_{\text{dx}}$ and $s_{\text{dy}}$. This reduces the map size by a factor of $s_{\text{dx}}\times s_{\text{dy}}$ while preserving semantic content through a compact 2-D backbone followed by layer normalization, yielding $X_{\text{sc}}$ in \myeq{eq:Xsc}. We then project each 3-D token onto this planar grid, round its coordinates, and collect the unique cell indices $(u_j,v_j)$ defined in \myeq{eq:uv}. By processing only occupied cells, we eliminate spatial duplicates and ensure each frequency transform is computed once.

For every retained cell, we crop a square window $P_j$ from $X_{\text{sc}}$ as in \myeq{eq:patch}. We choose the window size $H_k\times W_k$ (empirically set to $16\times16$) to cover a vehicle’s immediate neighborhood while maintaining a small FFT kernel. An orthonormal 2D DFT applied to each window produces the complex spectrum $\widehat{P}_j$ in \myeq{eq:dft}. Because we compute the DFT per occupied cell rather than over the full map, computational complexity scales with the number of occupied cells.

From the magnitude spectrum, we derive four statistics:
(i) the Direct Current (DC) component, or zero-frequency component (\myeq{eq:dc}), which quantifies average activation in the local neighborhood;
(ii) low-frequency energy $E_j^{\text{low}}$ (\myeq{eq:low}), capturing slowly varying background patterns;
(iii) high-frequency energy $E_j^{\text{high}}$ (\myeq{eq:hi}), measured outside radius $\beta k$ to emphasize sharp edges and fine details;
(iv) their ratio $S_j$ (\myeq{eq:S}), which serves as an edge saliency score highlighting vehicle contours.
In our experiments, we fix the band-split parameters to $\alpha=0.125$ and $\beta=0.25$. We concatenate these four statistics into a $4d$-dimensional vector $\nu_j$ (\myeq{eq:phi}), stack all $\nu_j$ into the matrix $\nu_{\text{freq}}$ (\myeq{eq:phifreq}), and apply a linear projection to produce the frequency-aware token representation $Q_{\text{freq}}$ (\myeq{eq:proj}), which matches the feature dimensionality of the $C$ matrix in the FSSM.
 This compact descriptor enriches sparse 3D tokens with local spectral cues while keeping both memory footprint and runtime modest.

\subsection{Ablation on downsampling rate for frequency feature extraction}


To minimize computational overhead, we first downsample the global scene features along their spatial dimensions and then extract frequency domain features. Reducing the feature map size effectively enlarges the grid, allowing each grid to represent a larger local area of the scene. Consequently, more point-level tokens are assigned to the same grid cell, significantly reducing the number of unique cells $(u_j,v_j)$. This, in turn, decreases the frequency of FFT computations, thereby lowering the overall computational cost.

To investigate the impact of the downsampling rate on model performance, we conduct comparative experiments as detailed in \mytab{table:freq-downsample}. The results indicate that higher downsampling rates lead to a decline in model performance. This is attributed to the substantial loss of high-frequency information—crucial for delineating object boundaries and enhancing localization accuracy—during the convolutional downsampling of the global scene feature map. Such frequency features are less beneficial for the FSSM, as they do not provide additional frequency domain cues. However, higher downsampling rates do offer reduced computational overhead. To strike a balance between performance and computational cost, we selected a downsampling rate of 4 for this study.

\begin{table}[!ht]
  \centering
  \caption{Performance of different frequency downsampling rates.}
  \label{table:freq-downsample}
  \small
  \begin{tabular}{@{}l c c@{}}
    \toprule
        \multirow{2}{*}{Downsampling rate}
      & \multicolumn{1}{c}{OPV2V}
      & \multicolumn{1}{c}{V2V4Real}      \\
      & AP@0.5/0.7   & AP@0.5/0.7                      \\
    \midrule
    1                & 0.974 / 0.933     & 0.638 / 0.430                    \\
    2                & 0.976 / 0.931     & 0.642 / 0.446                    \\
    \rowcolor[HTML]{EFEFEF}
    4--\ourname      & 0.973 / 0.934     & 0.644 / 0.447                    \\
    8                & 0.963 / 0.925     & 0.624 / 0.410                  \\
    16               & 0.965 / 0.922     & 0.576 / 0.357                   \\
    \bottomrule
  \end{tabular}
\end{table}

\begin{figure*}[!ht]
    \centering
    \begin{subfigure}[t]{0.33\textwidth}
        \centering
        \includegraphics[width=\linewidth]{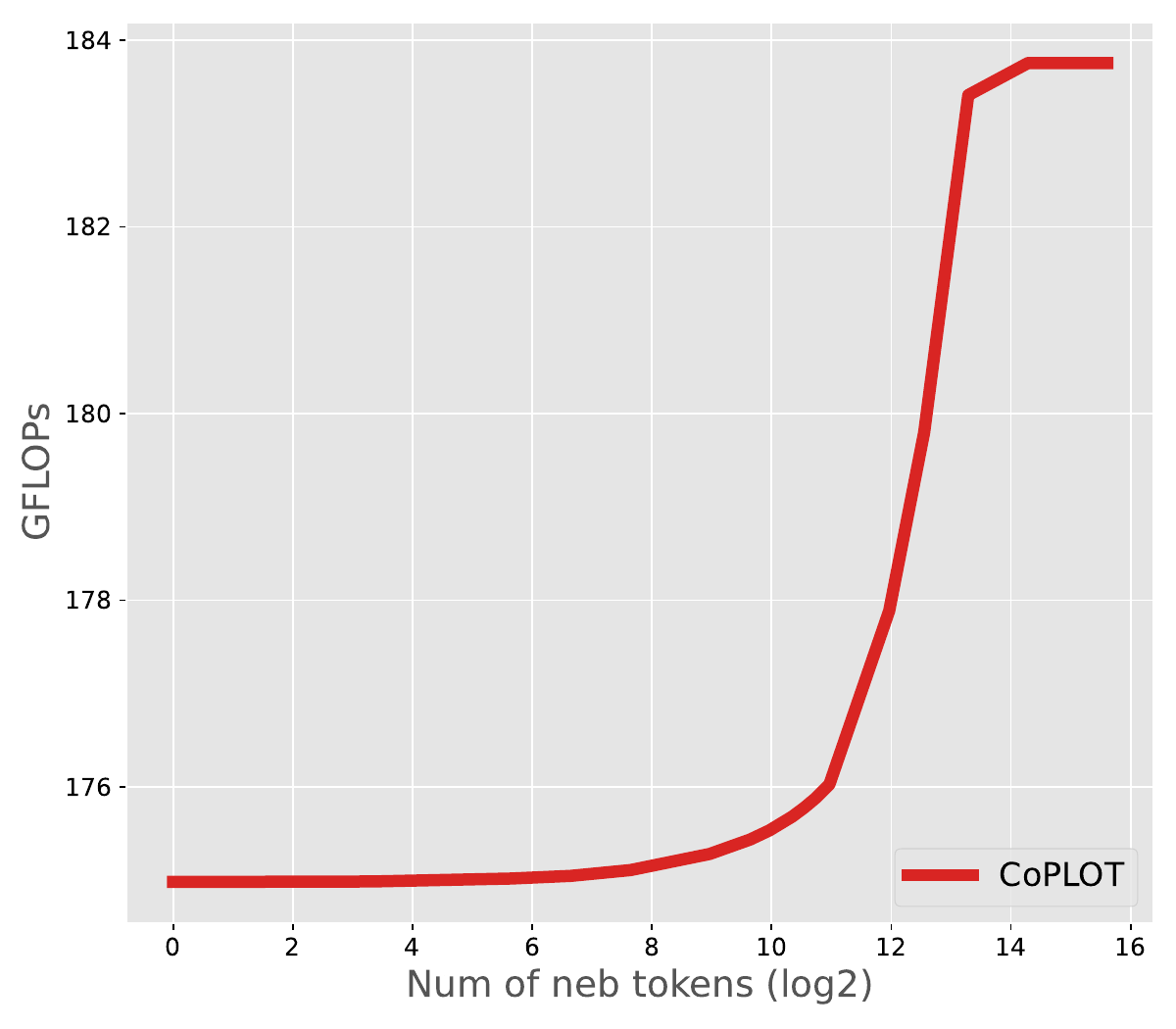}
    \end{subfigure}
    \begin{subfigure}[t]{0.33\textwidth}
        \centering
        \includegraphics[width=\linewidth]{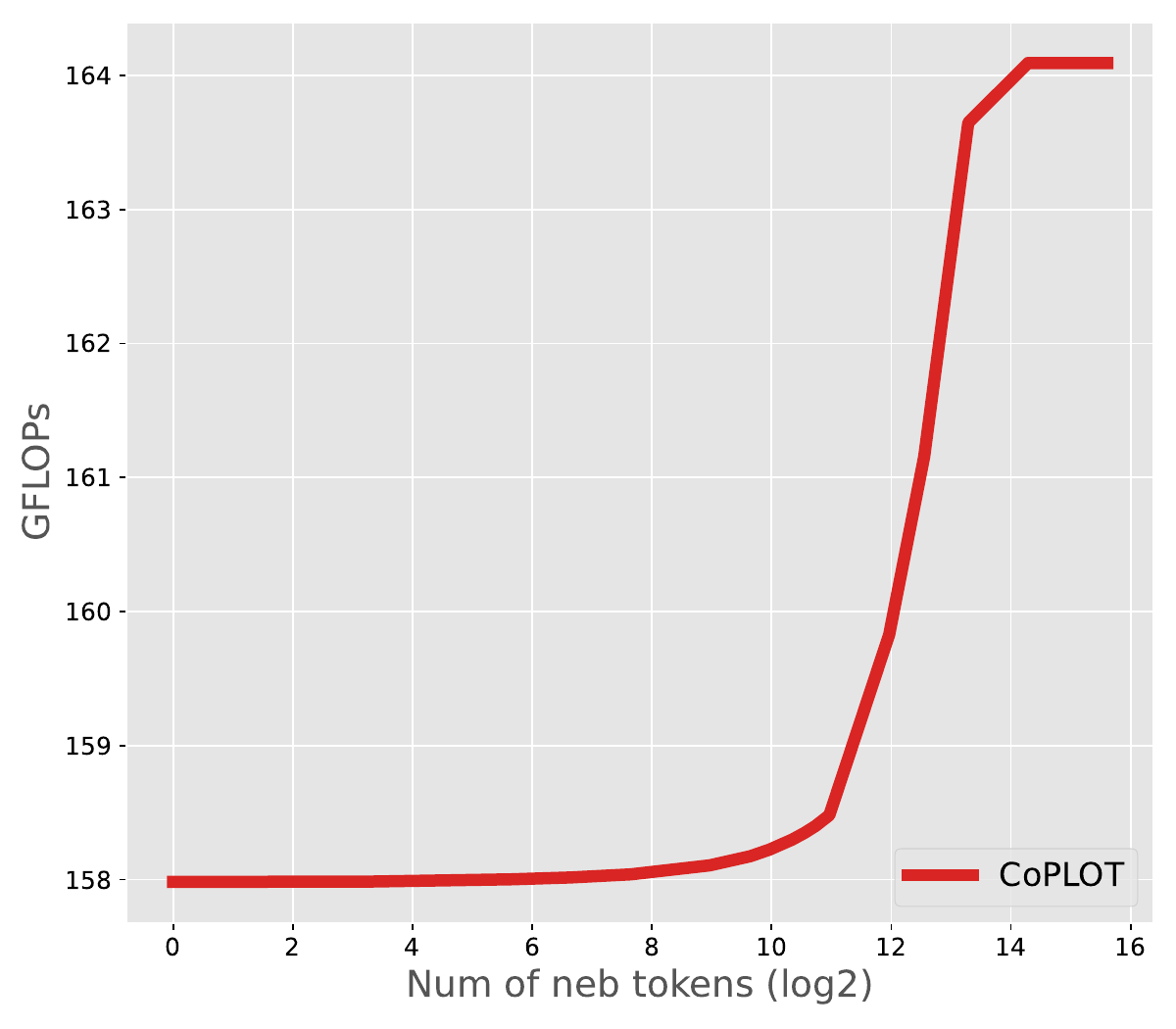}
    \end{subfigure}
    \begin{subfigure}[t]{0.33\textwidth}
        \centering
        \includegraphics[width=\linewidth]{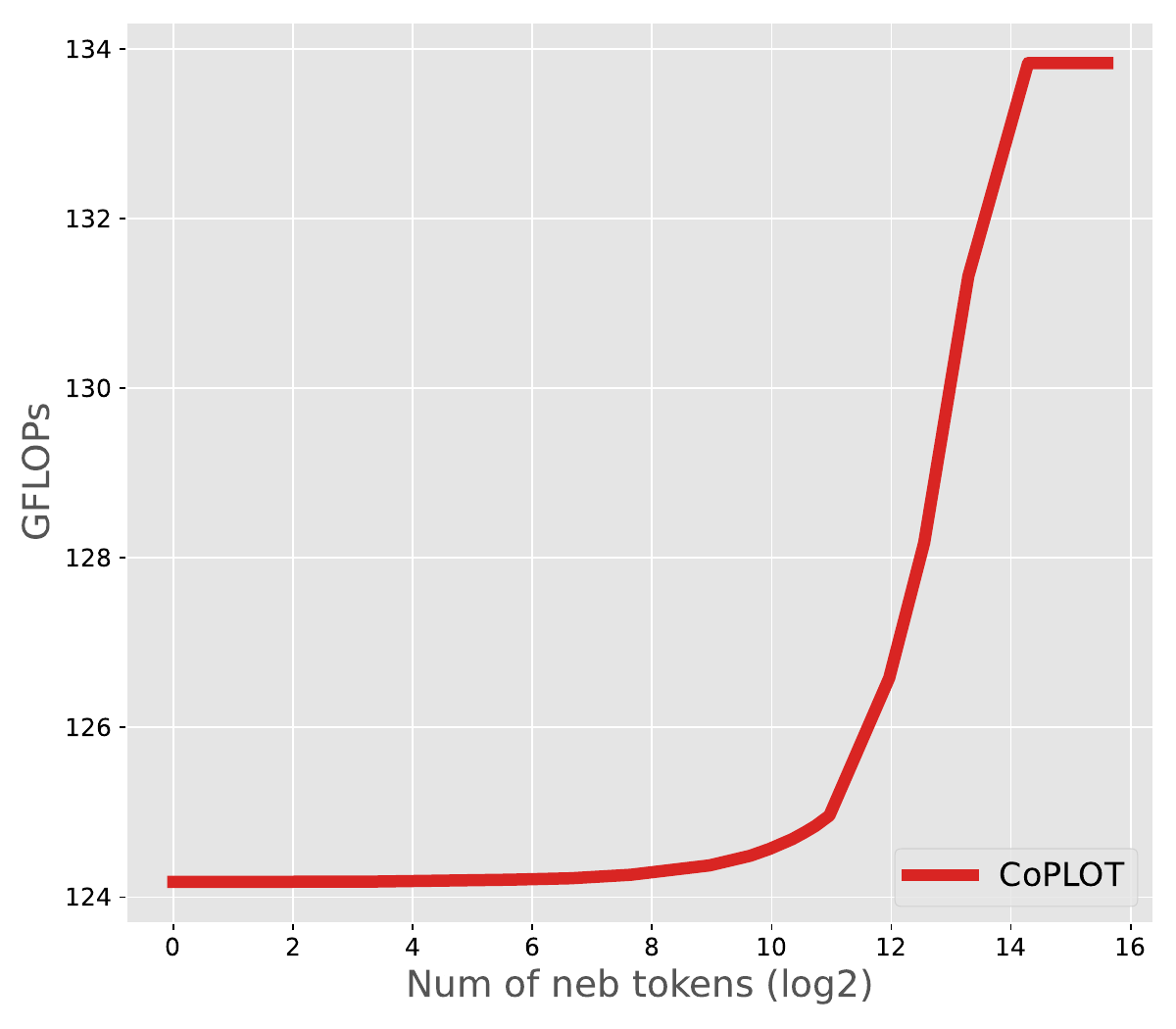}
    \end{subfigure}
    \caption{Relationship between the number of point-level tokens selected by neighbor agents and computational overhead in the \ourname framework. The x-axis indicates the number of selected tokens in log$_2$ scale.}
    \label{fig:ntokenflopsaa}

\end{figure*}

\begin{figure*}[!ht]
    \centering
    \begin{subfigure}[t]{0.33\textwidth}
        \centering
        \includegraphics[width=\linewidth]{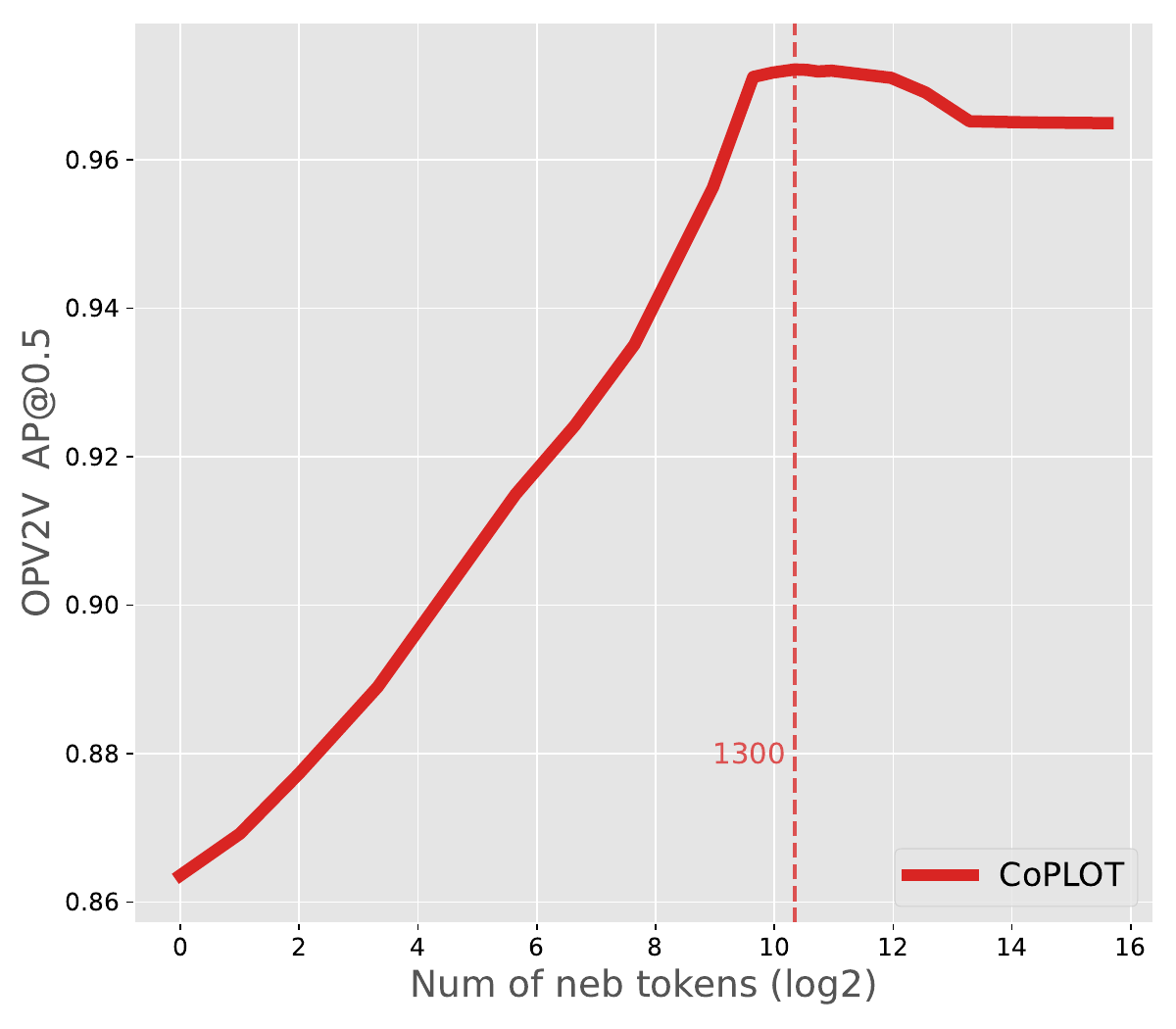}
    \end{subfigure}
    \begin{subfigure}[t]{0.33\textwidth}
        \centering
        \includegraphics[width=\linewidth]{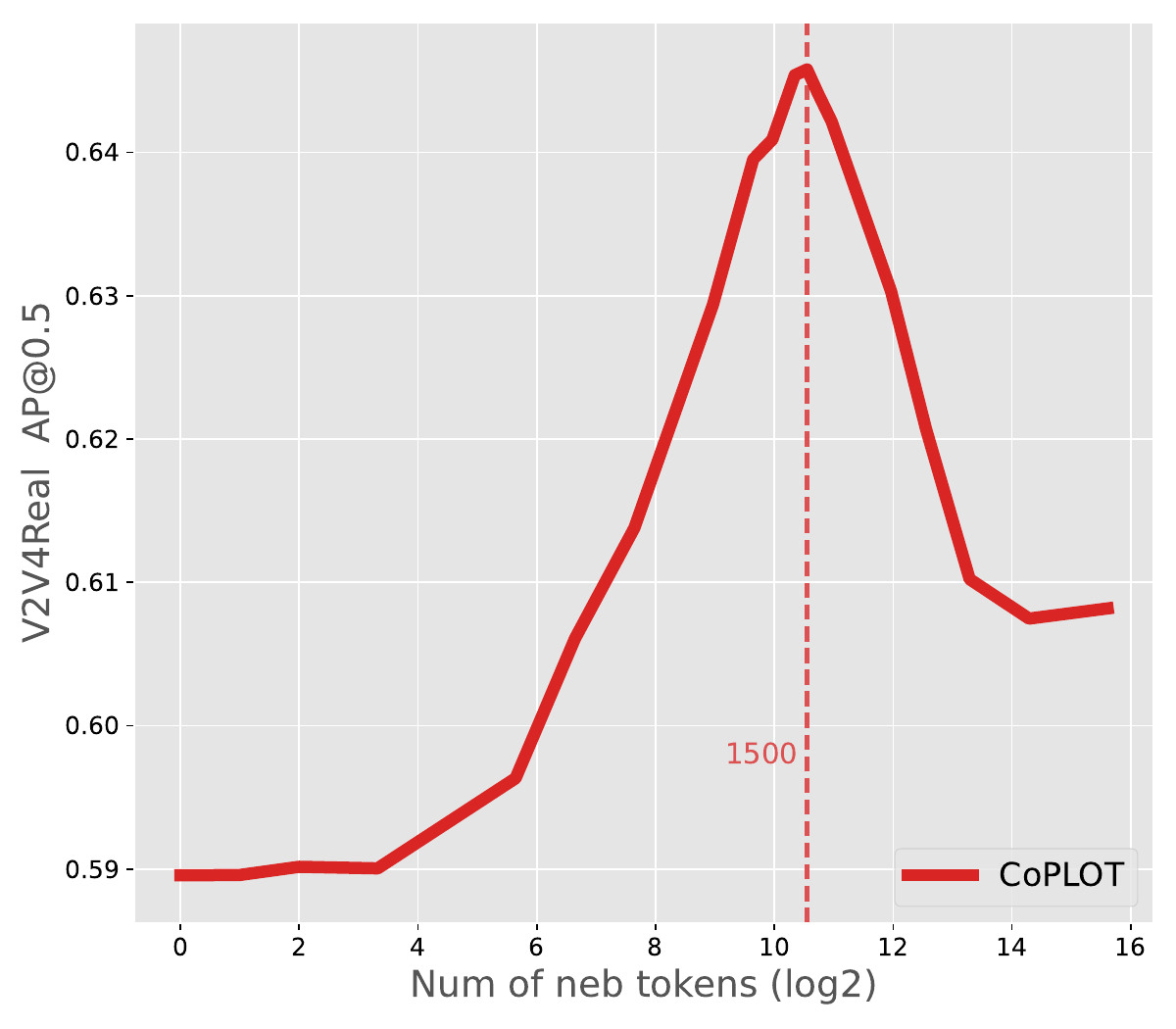}
    \end{subfigure}
    \begin{subfigure}[t]{0.33\textwidth}
        \centering
        \includegraphics[width=\linewidth]{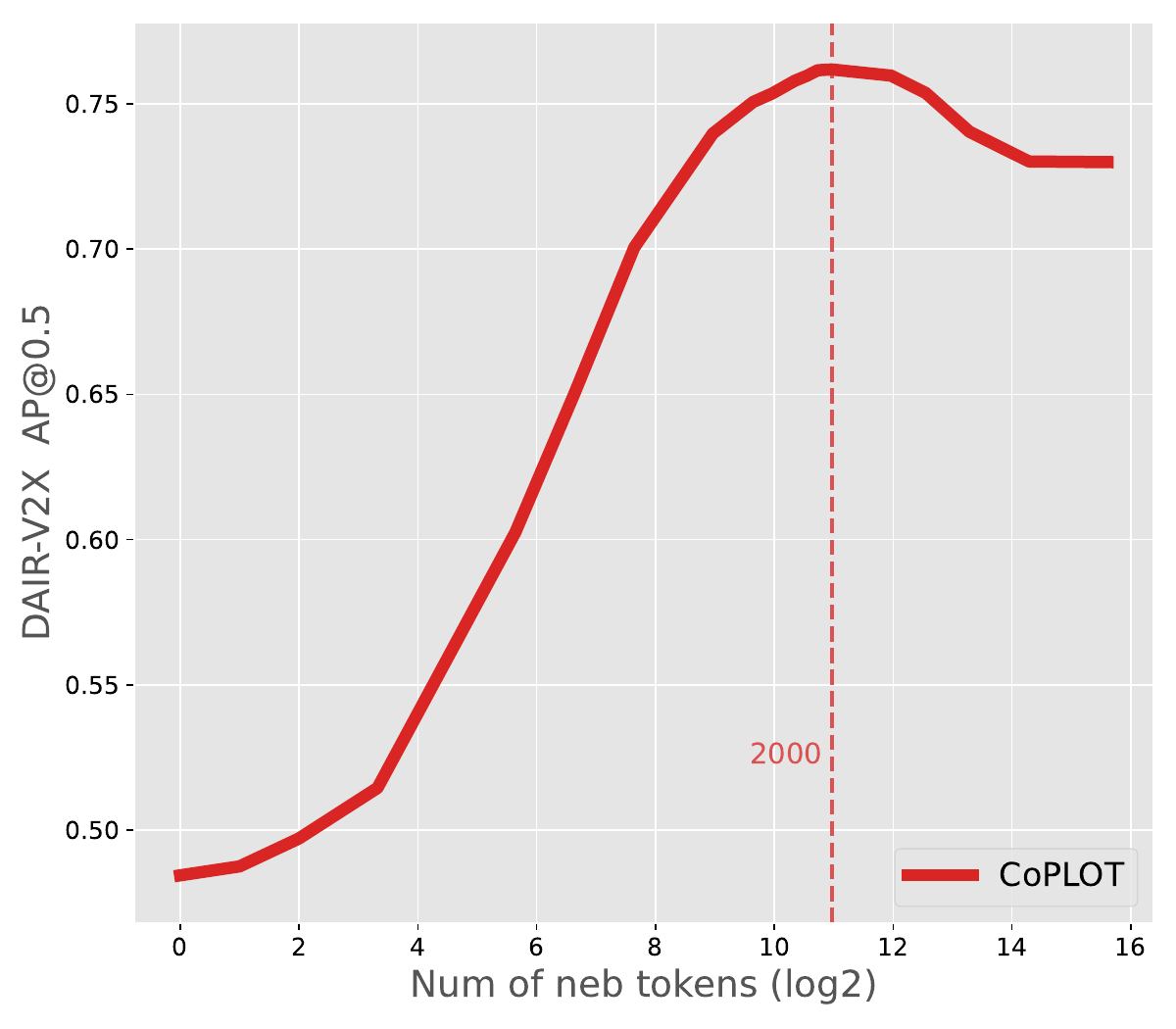}
    \end{subfigure}
    \caption{Relationship between the number of point-level tokens selected by neighbor agents and collaboration performance in the \ourname framework. The x-axis indicates the number of selected tokens in log$_2$ scale.}
    \label{fig:ntokenapaaa}

\end{figure*}

\section{Additional Experiments and Analysis}

\subsection{Computational overhead for different neb-token counts}

To more clearly illustrate the relationship between the number of top-$k$ tokens selected from neighboring agents and computational overhead, we plot this relationship across three datasets in \myfig{fig:ntokenflopsaa}. The curves follow an exponential trend; however, since the horizontal axis represents the base-2 logarithm of the top-$k$ token count, this corresponds to a linear increase in overhead with the number of selected tokens. This finding confirms that \ourname’s computational cost grows linearly with the length of the point-level token sequence, in contrast to the quadratic scaling observed in transformer or CNN models.

\subsection{Perception performance for different neb-token counts}

In the preceding figures (\myfig{fig:compt50} and \myfig{fig:communication_dairv2x_ap_50}), we observed that perception accuracy initially improves with greater computational or communication investment but then peaks and declines slightly. To analyze this behavior, we plot accuracy versus the number of top-$k$ tokens selected from neighboring agents in \myfig{fig:ntokenapaaa}. The curve rises at first, reaches a maximum, and then falls off modestly as $k$ increases.

\textbf{What Causes the Performance Peak?} 

During training, we vary each neighbor agent’s top-$k$ token threshold uniformly between 2 and 2000, randomly sampling a value for each agent in every batch. This range encourages the model to fuse neighbor-token sequences of diverse lengths effectively. At test time, we fix $k$ to specific values and measure the resulting performance. 

During the training stage, the relatively small maximum $k$ was chosen to promote the transmission of compact point-level token sequences, thereby reducing both communication and computational overhead. However, this training regimen \textit{introduces an inductive bias}: when tested with token counts beyond the training maximum, the model must process additional background or noisy tokens, which imposes extra burden and leads to a slight drop in accuracy. In the two real-world datasets shown in \myfig{fig:ntokenapaaa}, this decline after the peak is more pronounced, reflecting their greater scene complexity and feature variability. These results also indicate that once the model receives a sufficiently compact yet comprehensive token set, additional shared tokens offer no further accuracy gains, revealing an upper bound on the utility of neighbor-shared information.

\textbf{Why Does the Peak Occur at This Token Count? }

Taking the OPV2V dataset as an example, we investigate why the perception performance peaks when each neighboring agent shares approximately 1300 tokens. By analyzing the test set, we find that the average scene contains roughly 15 foreground vehicles (with a maximum of 33), each having average dimensions of approximately $4.5\times2.0\times1.6\,\text{m}$. The average lateral area of each vehicle is thus estimated as $(4.5+2.0)\times2\times1.6=20.8\,\text{m}^2$. Since LiDAR points typically project onto only part of a vehicle’s visible surfaces—more area when closer and less when farther away—we approximate the average projected area to be two-thirds of the total lateral area, resulting in approximately $20.8\times\frac{2}{3}=13.867\,\text{m}^2$.

Given our grid interval for point-level tokens is $0.4\,\text{m}$, each vehicle requires approximately $13.867/(0.4\times0.4)\approx86.67$ tokens to represent its projected area. Considering an average of 15 vehicles per scene, the total number of tokens needed to adequately represent the object-related foreground information is roughly $86.67\times15\approx1300$, matching precisely the experimentally observed performance peak in \myfig{fig:ntokenapaaa}. A similar analysis can be performed for the other datasets.

However, it is important to clarify that this calculation is based on average statistics from the entire test dataset. Individual scenes vary significantly in complexity and vehicle density, so the optimal number of tokens to share may differ from scene to scene. This highlights the need for a dynamic method that can adaptively select the top-$k$ threshold based on the complexity of each specific scene and task, thereby ensuring higher reliability and better adaptability to real-world scenarios. Moreover, the exact optimal value fluctuates slightly around 1300 due to the relatively coarse sampling intervals used in testing, leaving room for further refinement.

Overall, our exploration of computational cost, communication overhead, and perception accuracy confirms that these factors can be jointly optimized. This represents a step beyond communication-only methods like Where2comm and offers greater adaptability for real-world deployment.

\subsection{Comparison of different reordering methods}

In the ablation study presented in \mytab{table:ablation}, we demonstrate that removing the semantic-aware token reordering module significantly reduces model performance, highlighting the substantial positive contribution of our proposed \ourname. The semantic-aware reordering module provides a robust foundation for subsequent FSSM modeling by establishing a semantically coherent token sequence.

To further validate the effectiveness of our semantic-aware token reordering approach, we conducted additional comparative experiments summarized in \mytab{table:reordering-methods}. Specifically, we compared our method with several commonly used reordering techniques, including raster scan, Z-order curve, Hilbert curve, and random reordering as a baseline. The experimental results clearly show that our approach outperforms these alternatives by a significant margin. This advantage arises from the module’s capability to dynamically reorder tokens based on both global scene context and the semantic characteristics of each individual point-level token, thereby maintaining semantic coherence and spatial proximity even after reordering.

In contrast, the baseline methods utilize fixed ordering strategies or templates that do not account for the dynamic nature of scene context or token semantics. Consequently, their resulting token sequences are not optimally suited for the subsequent FSSM processing. Notably, even the relatively advanced Hilbert curve ordering fails to surpass the simpler Z-order curve in perception accuracy, suggesting that the effectiveness of a serialization method can vary significantly depending on the specific task and scene context.
Furthermore, in our semantic-aware token reordering module, tokens sharing the same semantic group initially follow a predefined Z-order curve during the early training stage. This design choice preserves local spatial proximity and reduces unnecessary randomness, assisting downstream modules in establishing stable semantic representations, particularly in the early stages of training, while still maintaining high flexibility and adaptability in token sequencing.

\begin{table}[!ht]
  \centering
  \caption{Performance of different reordering methods.}
  \label{table:reordering-methods}
  \small
  \begin{tabular}{@{}l c@{}}
    \toprule
    Reordering Method & OPV2V AP@0.5/0.7 \\
    \midrule
    Random            & 0.942 / 0.871     \\
    Raster scan       & 0.955 / 0.901     \\
    Z-order curve     & 0.972 / 0.911     \\
    Hilbert curve     & 0.971 / 0.910     \\
    \rowcolor[HTML]{EFEFEF}
    \ourname             & 0.973 / 0.934     \\
    \bottomrule
  \end{tabular}
\end{table}

\subsection{Comparison of different semantic group number}

In our semantic-aware token reordering module, each point-level token is  assigned a semantic group index, ensuring that tokens with identical indices are placed closer together in the final one-dimensional sequence. Here, the semantic group number is a manually defined hyperparameter. In the experiments summarized in \mytab{table:semantic-group-num}, we investigate how varying the semantic group number affects model performance. Experimental results reveal that setting a small semantic group number limits model performance, primarily because coarse semantic grouping causes tokens sharing similar semantic characteristics to remain relatively distant even within the same group. Conversely, setting a large semantic group number requires extensive training to achieve optimal performance to some extent and introduces additional computational overhead. Balancing performance and training cost, we select 48 as the semantic group number for our framework. It is important to clarify that this choice is based on statistical analysis and empirical observation, and thus may not be universally optimal across all datasets or scenarios. A potentially superior solution would involve dynamically adjusting the semantic group number according to scene complexity, dynamics, and collaboration requirements.

\begin{table}[!ht]
  \centering
  \caption{Effect of semantic group number on perception performance.}
  \label{table:semantic-group-num}
  \small
  \begin{tabular}{@{}c c c@{}}
    \toprule
    \multirow{2}{*}{Semantic group num}
      & \multicolumn{1}{c}{OPV2V}
      & \multicolumn{1}{c}{V2V4Real}      \\
      & AP@0.5/0.7   & AP@0.5/0.7                      \\
    \midrule
    2    & 0.963 / 0.926 & 0.619 / 0.418 \\
    8    & 0.968 / 0.929 & 0.593 / 0.403 \\
    16   & 0.968 / 0.930 & 0.634 / 0.423 \\
    32   & 0.970 / 0.928 & 0.646 / 0.443 \\
    \rowcolor[HTML]{EFEFEF}
    48--\ourname    & 0.973 / 0.934 & 0.644 / 0.447 \\
    64   & 0.973 / 0.931 & 0.579 / 0.395 \\
    128  & 0.971 / 0.907 & 0.569 / 0.381 \\
    \bottomrule
  \end{tabular}
\end{table}

\subsection{Comparison of different encoder methods}

In this study, considering the limited receptive fields and poor scalability of convolution-based methods and the quadratic computational complexity of transformers—which impedes processing long sequences—we select emerging state-space models (SSMs) as our primary encoding approach, due to their linear computational complexity and comparable performance to transformers. We enhance conventional SSMs to adapt them specifically for our collaborative perception task based on point-level tokens.

To further examine the impact of different encoder methods on model performance, we conduct comparative experiments presented in \mytab{table:encoder-method-comparison}. We compare our proposed method against Sparse Conv, Point Transformer V3 (PTV3) \cite{wuPointTransformerV32024}, and Linear. In the table, the number following "PTV3" indicates the attention window size—for instance, "PTV3-1024" represents an attention window size of 1024.
Experimental results show that our proposed \ourname framework outperforms competing methods in perception accuracy while demonstrating significant advantages in model efficiency, benefiting from our tailored pipeline designed specifically for optimized point-level tokens. Particularly, PTV3 incurs prohibitively large computational costs when processing longer sequences, making it unsuitable for practical deployment to a certain extent. Additionally, smaller-window PTV3 variants perform worse than larger-window counterparts, highlighting the importance of capturing long-range spatial structures and semantic dependencies for accurate and expressive token representations.
In summary, the experimental findings in \mytab{table:encoder-method-comparison} reaffirm that our proposed \ourname achieves an effective balance between perception performance and computational efficiency.


\begin{table}[!ht]
  \centering
  \caption{Performance of different encoder methods.}
  \label{table:encoder-method-comparison}
  \small
  \begin{tabular}{@{}l c c c c@{}}
    \toprule
    \shortstack{Encoder \\ Method}   & \shortstack{OPV2V\\AP@0.5/0.7} & GFLOPs            & \shortstack{Peak GPU\\Mem (GB)} & Params (M) \\
    \midrule
    Linear           & 0.943 / 0.869                  & 162.403          & 0.3385            & 7.351      \\
    Sparse Conv      & 0.954 / 0.897                  & 179.364          & 0.3643            & 13.769     \\
    PTV3-256         & 0.967 / 0.923                  & 174.197          & 0.5822            & 7.312      \\
    PTV3-1024        & 0.968 / 0.930                  & 185.959          & 0.9991            & 7.312      \\
    PTV3-4096        & 0.971 / 0.933                  & 236.106          & 4.1601            & 7.312      \\
    \rowcolor[HTML]{EFEFEF}
    \ourname             & 0.973 / 0.934                  & 176.210          & 0.5780            & 6.368      \\
    \bottomrule
  \end{tabular}
\end{table}

\subsection{Comparison of different scope combinations in the FSSM}

In our frequency-enhanced state-space model (FSSM), following prior work, we adopt a dual-scope modeling strategy. Specifically, in the global scope, the FSSM processes the entire token sequence to capture long-range spatial dependencies, while in the local scope, it operates within smaller windows to model fine-grained local spatial structures. The combination of these two scopes produces richer and more comprehensive feature representations.

In \mytab{table:fssm-scope}, we investigate the impact of different scope settings on model performance, where ``$\infty$" indicates feeding the entire point-level token sequence directly into the FSSM without partitioning. Results demonstrate that dual-scope modeling outperforms single-scope modeling, as well as capturing long-range dependencies improves the learned feature representations. Furthermore, combining long-range dependencies with short-range modeling helps yield enriched feature representations. In the collaborative 3D detection task addressed in this work, this approach effectively models the spatial relationships between vehicles, between vehicles and background, and provides more comprehensive semantic context.

\begin{table}[!ht]
  \centering
  \caption{Comparison of different scope combinations in the FSSM.}
  \label{table:fssm-scope}
  \small
  \begin{tabular}{@{}c c c@{}}
    \toprule
    \multirow{2}{*}{Scope Combination}
      & \multicolumn{1}{c}{OPV2V}
      & \multicolumn{1}{c}{V2V4Real}        \\
      & AP@0.5/0.7   & AP@0.5/0.7                   \\
    \midrule
    $[128]$           & 0.952 / 0.913    & 0.587 / 0.371        \\
    $[1024,128]$      & 0.957 / 0.915    & 0.599 / 0.397        \\
    $[2048,128]$      & 0.934 / 0.917    & 0.588 / 0.321        \\
    $[\infty,64]$     & 0.963 / 0.913    & 0.550 / 0.375        \\
    \rowcolor[HTML]{EFEFEF}
    $[\infty,128]$    & 0.973 / 0.934    & 0.644 / 0.447        \\
    $[\infty,256]$    & 0.966 / 0.924    & 0.639 / 0.443        \\
    $[\infty,512]$    & 0.958 / 0.929    & 0.582 / 0.418        \\
    $[\infty]$        & 0.969 / 0.926    & 0.602 / 0.394        \\
    \bottomrule
  \end{tabular}
\end{table}

\subsection{Comparison of different single-agent perception methods}

Extensive comparative experiments have demonstrated the effectiveness of our proposed \ourname framework for multi-agent collaborative perception. It achieves an excellent balance among perception accuracy, computational overhead, and communication efficiency. Additionally, our point-level token representation effectively overcomes intrinsic limitations associated with traditional 2D BEV features. To further evaluate the versatility of \ourname, we conduct a comparison against widely-used single-agent perception methods, specifically Point Pillar \cite{langPointPillarsFastEncoders2019} and Voxel Net \cite{zhouVoxelNetEndtoendLearning2018}.
\mytab{table:pointpillar-voxel-ours} compares single-agent perception accuracy across three datasets, clearly showing that \ourname outperforms the baselines. Point Pillar exhibits comparatively lower accuracy because its pipeline compresses vertical information early, causing substantial loss of spatial structural detail. Voxel Net leverages 3D convolutions within voxelized space, effectively preserving spatial structure, yet has limited receptive fields and thus inferior capability to model long-range spatial dependencies compared to our \ourname, leading to slightly weaker performance.
Moreover, \mytab{table:computation-memory} presents computational efficiency metrics. Although Voxel Net achieves competitive performance, it requires computationally intensive multi-layer 3D convolutions, resulting in significantly higher GFLOPs, GPU memory usage and parameter counts. In contrast, our \ourname exhibits superior computational efficiency, achieving substantial advantages in GFLOPs and model size. These experimental results indicate that \ourname can effectively balance perception accuracy and computational overhead  in single-agent scenarios, further highlighting its advanced designs.

\begin{table}[!ht]
  \centering
  \caption{Comparison of different single-agent perception methods across three datasets.}
  \label{table:pointpillar-voxel-ours}
  \small 
  \begin{tabular}{@{}l c c c@{}}
    \toprule
    \multirow{2}{*}{Method}
      & \multicolumn{1}{c}{OPV2V}
      & \multicolumn{1}{c}{V2V4Real}
      & \multicolumn{1}{c}{DAIR-V2X} \\ 
    & {AP@0.5/0.7} & {AP@0.5/0.7} & {AP@0.5/0.7} \\
    \midrule
    Point Pillar  & 0.898 / 0.747    & 0.611 / 0.407    & 0.646 / 0.546 \\
    Voxel Net     & 0.902 / 0.843    & 0.712 / 0.477    & 0.804 / 0.673 \\
    \rowcolor[HTML]{EFEFEF}
    \ourname           & 0.905 / 0.844    & 0.713 / 0.481    & 0.801 / 0.688 \\
    \bottomrule
  \end{tabular}
\end{table}

\begin{table}[!ht]
  \centering
  \caption{Comparison of model efficiency across single-agent perception methods on OPV2V.}
  \label{table:computation-memory}
  \small
  \begin{tabular}{@{}l c c c@{}}
    \toprule
    Method        & GFLOPs              & Peak GPU Mem (GB)  & Params (M) \\
    \midrule
    Point Pillar  & 151.216    & 0.263       & 8.058      \\
    Voxel Net     & 188.352    & 0.984      & 13.497     \\
    \rowcolor[HTML]{EFEFEF}
    \ourname      & 114.079    & 0.338       & 4.540      \\
    \bottomrule
  \end{tabular}
\end{table}

%


\fi

\end{document}